\documentclass[12pt]{report}
\usepackage[a4paper, margin=1in]{geometry}
\usepackage[parfill]{parskip}
\usepackage{graphicx}
\usepackage{setspace}
\usepackage{tocbibind}
\usepackage{titlesec}
\usepackage{hyperref}
\usepackage{algorithm}
\usepackage{amsmath}
\usepackage{algpseudocode}
\usepackage{amssymb}
\usepackage{booktabs}
\usepackage{mathbbol}
\usepackage{array}
\usepackage[section]{placeins}
\newcolumntype{C}{>{\centering\arraybackslash}m{.5in}}
\usepackage[numbers,sort&compress]{natbib}
\usepackage{arydshln}
\usepackage{afterpage}

\providecommand{\keywords}[1]
{
  \textbf{\textit{Keywords---}} #1
}

\hypersetup{
    colorlinks=true,
    linkcolor=black,
    citecolor=black,
    filecolor=black,
    urlcolor=black,
    pdfstartview=FitH
}

\usepackage[skip=10pt]{caption}

\graphicspath{ {./images/} }
\algnewcommand{\LineComment}[1]{\State  \(\triangleright\) #1 \hfill~}

\begin{document}
\pagenumbering{gobble} 
\begin{titlepage}
    \centering
    {\Large A Novel Approach To Implementing\\ Knowledge Distillation In Tsetlin~Machines}\par
    \vspace{1cm}
    {Calvin John Kinateder}\par
    {\small B.S. Computer Science}\par
    {M.S. Computer Science}\par
    {Department of Computer Science, University of Cincinnati}\par
    {\today}\par
    \null\vfill
    \includegraphics[width=0.5\textwidth\centering]{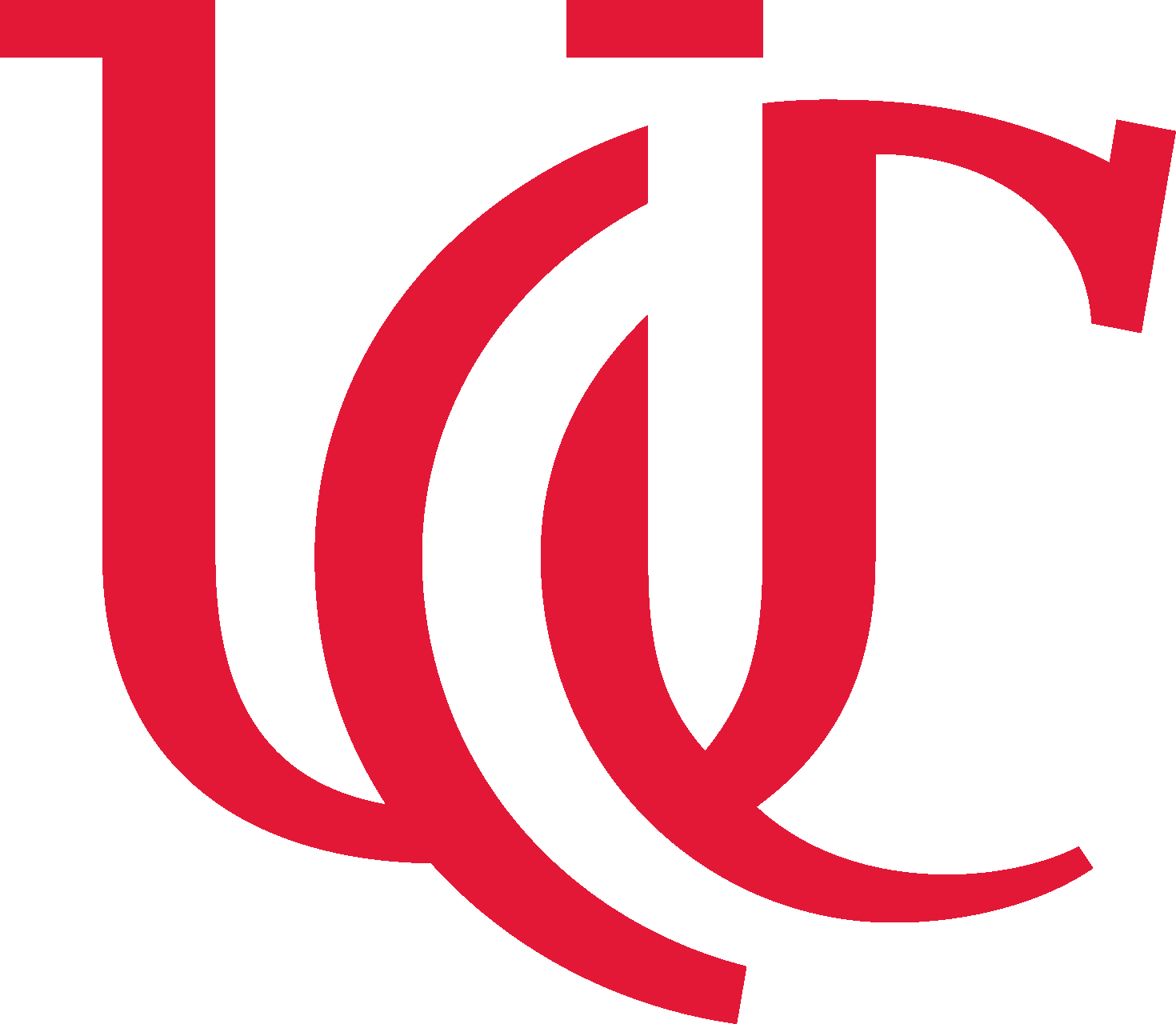}
    \vfill\null
    Committee Chair: Dr. Justin Zhan\par
\end{titlepage}

\clearpage
\pagenumbering{roman}

\chapter*{Abstract}
The Tsetlin Machine (TM) is a propositional logic based model that uses conjunctive clauses to learn patterns from data. As with typical neural networks, the performance of a Tsetlin Machine is largely dependent on its parameter count, with a larger number of parameters producing higher accuracy but slower execution. Knowledge distillation in neural networks transfers information from an already-trained teacher model to a smaller student model to increase accuracy in the student without increasing execution time. We propose a novel approach to implementing knowledge distillation in Tsetlin Machines by utilizing the probability distributions of each output sample in the teacher to provide additional context to the student. Additionally, we propose a novel clause-transfer algorithm that weighs the importance of each clause in the teacher and initializes the student with only the most essential data. We find that our algorithm can significantly improve performance in the student model without negatively impacting latency in the tested domains of image recognition and text classification.
\\
\\
\keywords{Tsetlin Machine, Knowledge Distillation, Explainable AI, Propositional Logic, Classification, Data Mining}
\clearpage

\newpage
\thispagestyle{empty}
\vspace*{\fill}
\begin{center}
    {© \the\year\ Calvin John Kinateder} \\[1em]
    {\today}
\end{center}
\newpage

\chapter*{Preface}
This document is the culmination of my research performed through my master’s program at the University of Cincinnati. I am incredibly grateful to the advisors and professors who have worked alongside me and helped me every step of the way. This endeavor would not have been possible without my research advisor, Dr. Usman Anjum. I would also like to thank Dr. Justin Zhan, my advisor and thesis committee chairman, for his advice and guidance. Additionally, I am most grateful for my committee members, Dr. John Gallagher and Dr. Vikram Ravindra, and their feedback and insight.
Finally, I would like to thank my family and friends for their support throughout my degree program.

\clearpage

\tableofcontents
\clearpage

\listoftables
\clearpage
\listoffigures
\clearpage

\chapter*{List of Symbols and Abbreviations}

\begin{tabular}{ll}
$TM$ & Tsetlin Machine \\
$KD$ & knowledge distillation \\
$C$ & set of TM clauses \\
$L$ & set of TM literals \\
$n$ & number of input features \\
$\zeta$ & number of output classes \\
$T$ & threshold TM parameter \\
$S$ & specificity TM parameter \\
$N$ & number of samples \\
$E$ & number of epochs \\
$K$ & number of times each experiment is aggregated over \\
$\alpha$ & ground truth/distribution balance \\
$\tau$ & distribution temperature \\
$z$ & weight transfer \\
$\delta$ & downsample rate \\
$\mathcal{T}'$ & training time \\
$\mathcal{T}$ & testing (inference) time \\
$Acc'$ & accuracy on training split \\
$Acc$ & accuracy on testing split \\
\end{tabular}

\clearpage
\pagenumbering{arabic}

\chapter{Introduction}
\section{Background}
\subsection{Knowledge Distillation}

\begin{figure}
    \centering
    \includegraphics[width=0.7\linewidth]{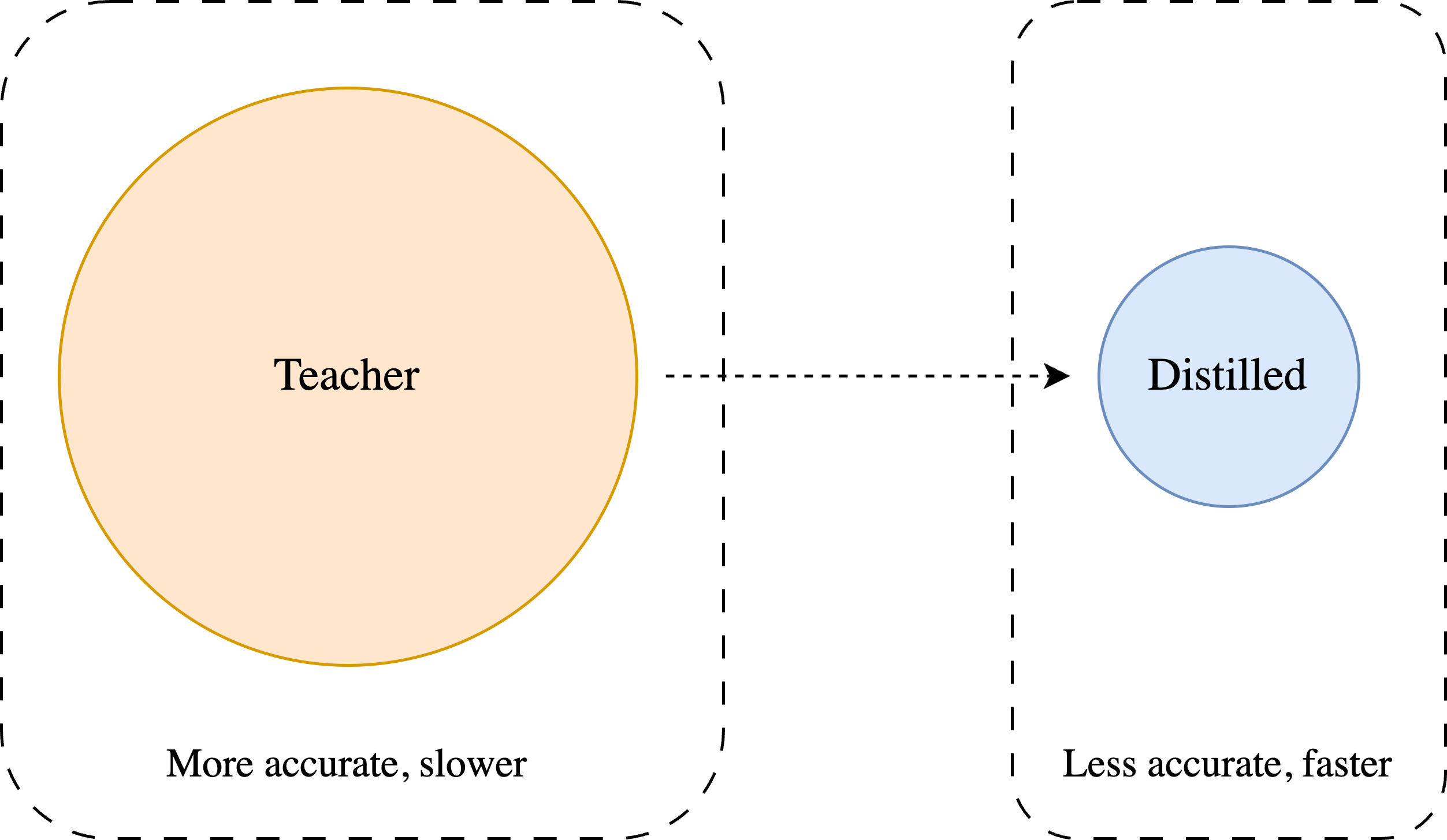}
    \caption{A visual representation of knowledge distillation.}
    \label{fig:kd-viz}
\end{figure}

Since their creation, deep neural networks have been incredibly prominent in solving a wide range of problems, including regression, object recognition, and text generation. As computing capability has advanced, so has the complexity of many neural networks, which has made such models both less explainable and more challenging to train. Additionally, complex models are slower in performance, which makes them harder to run on embedded hardware or in the field. Knowledge distillation, first introduced in 2015~\cite{hinton2015distillingknowledgeneuralnetwork} aims to ameliorate these problems by compressing a larger and more complex neural network into a simpler, smaller one while keeping similar performance. Figure~\ref{fig:kd-viz} illustrates the model size relationship.

\begin{figure}
    \centering
    \includegraphics[width=0.5\linewidth]{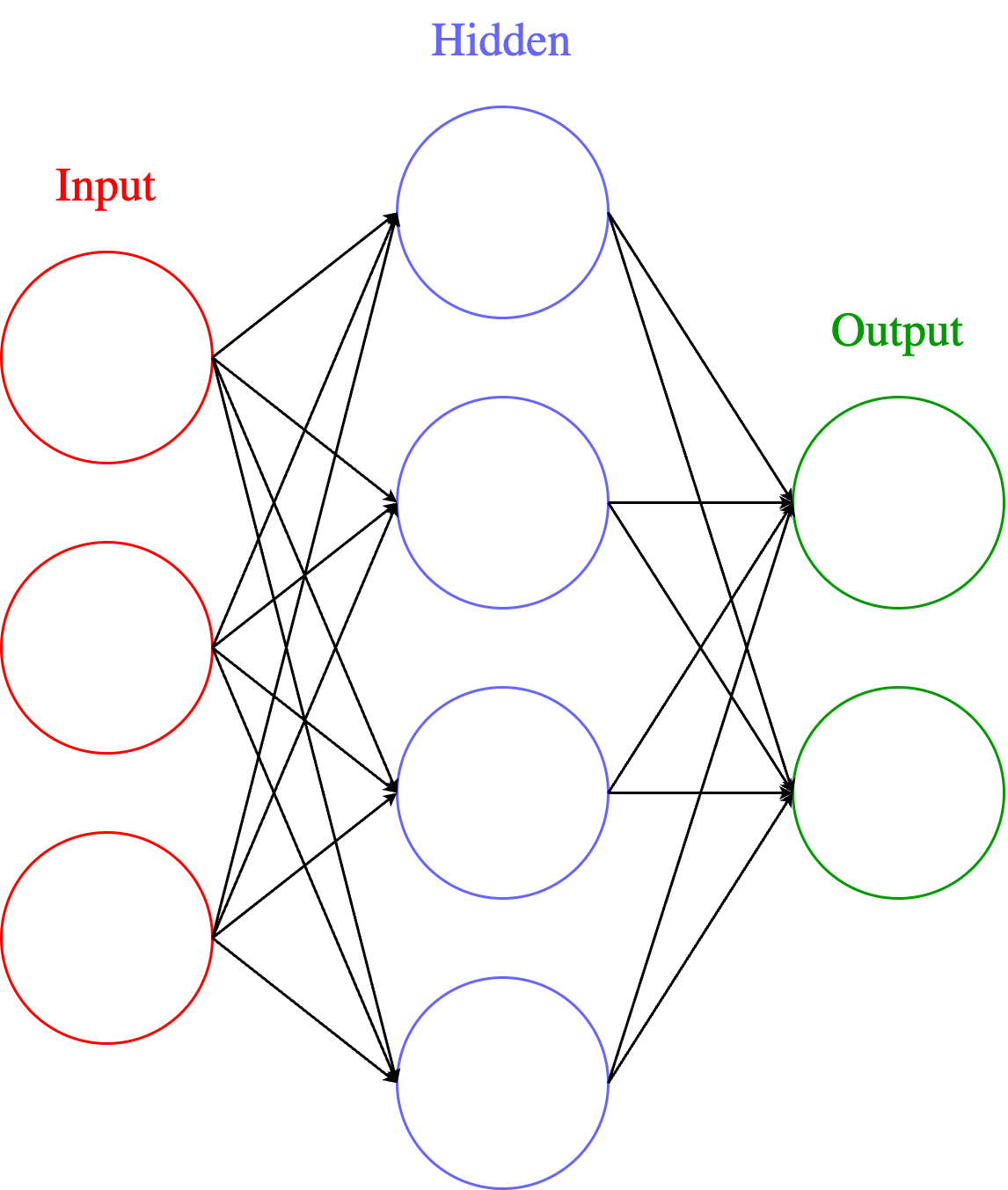}
    \caption{Typical layout of a neural network.}
    \label{fig:neural-network}
\end{figure}

Knowledge distillation is designed around a teacher-student model, where a smaller student model is taught to imitate the behavior of a larger teacher model. The goal is to have the smaller model take up less of a memory and computational footprint than the teacher model without a serious drop in performance. This idea is made possible by the internal structure of typical neural networks. A neural network is made up of many interconnected hidden layers (Figure~\ref{fig:neural-network}) that capture representations of the input data. These hidden representations can be viewed as a type of machinated ``knowledge'' that is ``learned'' during the training steps. In knowledge distillation, this hidden-layer knowledge in the teacher model is distilled and ``taught'' to the student through a supervised training process, where the student is rewarded for minimizing the difference between the inference of the teacher and the inference of the student over the training examples. This knowledge distillation is not just theorized, it is seen in practice~\cite{deepseekai2025deepseekr1incentivizingreasoningcapability}.

The goal of knowledge distillation is to teach a smaller model to mimic the behavior of a larger and more complex model. Some advantages of traditional knowledge distillation include:
\begin{itemize}
    \item Reduced memory footprint and computational requirements
    \item Quicker training and inference time when compared to the teacher
    \item Enhanced generalization
\end{itemize}
... while increasing the student's accuracy.

\begin{figure}
    \centering
    \includegraphics[width=0.67\linewidth]{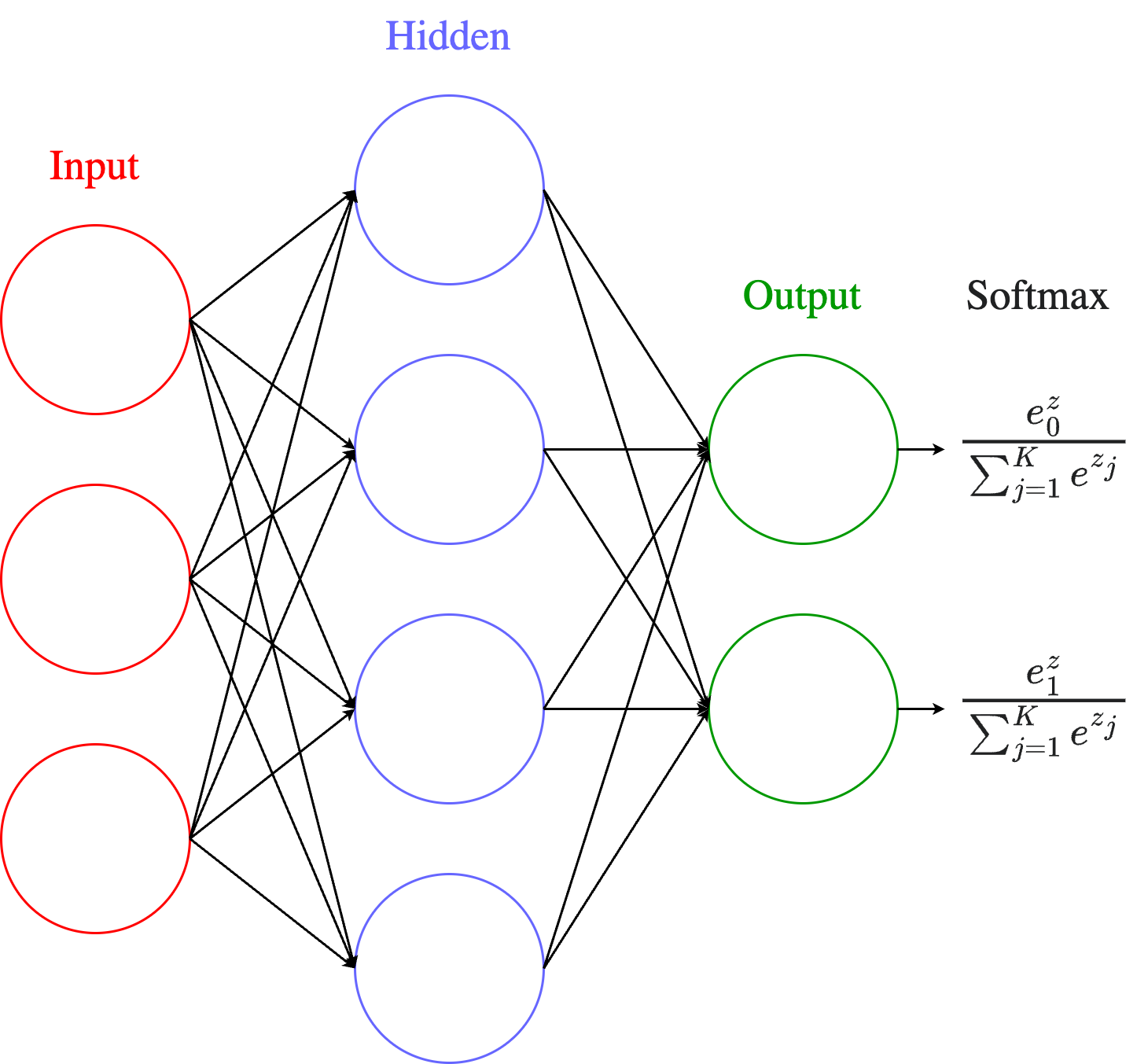}
    \caption{A typical neural network layout with softmax.}
    \label{fig:neural-network-w-softmax}
\end{figure}

There are two distinct steps in the knowledge distillation process: training the teacher model and training the student (or distilled) model. In typical knowledge distillation applications, the teacher model is trained in a normal fashion. After the teacher model has been trained, it can be used to obtain probabilistic ``soft'' labels for the training data. This can be done without significant effort. A standard neural network might have a final softmax~\cite{softmax} layer that takes an input array of real numbers and outputs it as a probability distribution. The neural network would then output the index with the highest number as its singular output. Equation~\eqref{eq:softmax} shows the softmax formula and Figure~\ref{fig:neural-network-w-softmax} shows the neural network with softmax applied.

\begin{equation} \label{eq:softmax}
	\sigma(z_i) = \displaystyle\frac{e^{z}_{i}}{\sum_{j=1}^K e^{z_{j}}} \ \ \ \text{for}\ i=1,2,\dots,K
\end{equation}

It follows that a model could be easily modified to output the array of probabilities instead of the index where the highest probability resides. The motivation for this approach is that there is more information contained in the probability array than a singular value. The student can then be trained to minimize the gap between its own softmax predictions and the teacher's softmax predictions. 

There are three different types of knowledge distillation methods, including:
\begin{itemize}
    \item Relation-based distillation~\cite{park2019relationalknowledgedistillation}
    \item Response-based distillation~\cite{yang2023categoriesresponsebasedfeaturebasedrelationbased}
    \item Feature-based distillation~\cite{Yang_Yu_Sheng_Yang_2023}
\end{itemize}

With relation-based knowledge distillation, the student model learns the relationship between the input data examples and the output labels. The teacher model creates a representation that expresses the relationships between the input data samples and the output labels in a matrix. Then, the student model's loss function is set up to minimize the difference between the relationship matrix generated by the teacher model and the ones predicted by the student model. This approach has the advantage of helping the student learn very advanced patterns that normally wouldn't be available for a smaller model to learn independently.

Response-based knowledge distillation (described above with the soft labels) is an outcome-driven task that focuses on minimizing the difference between the outputs of the last layer of both the teacher model and the student model. 
Response-based distillation is widely used in many domains and is the most common form of knowledge distillation~\cite{Ibm_2025}. It is the simplest form of knowledge distillation to implement, but is not always able to transfer as much knowledge as others.

Lastly, feature-based knowledge distillation trains the student model to imitate the internal, hidden features and representations that have already been learned and stored in the teacher model. Those internal representations are extracted from at least one hidden layer of the teacher model and used to help train the student model. This process is somewhat intensive. First, the teacher model is trained on the training examples in a standard fashion. Next, the student model is trained on the same features, while also minimizing the distance between its own hidden-layer representations and the hidden representations of the teacher model. An example metric here might be Kullback-Leibler Divergence, which measures the distance between two probability distributions~\cite{DBLP:journals/corr/Shlens14c}. Feature-based distillation can foster strong results, but only works when the structure of both the teacher and student model are closely similar.

In this paper, we will introduce a combination of feature- and response-based knowledge distillation.

There are three mainstream techniques for training student and teacher models:
\begin{itemize}
    \item Offline~\cite{Gou_2021}
    \item Online~\cite{Guo_2020_CVPR}
    \item Self-distillation~\cite{zhang2019teacherimproveperformanceconvolutional}
\end{itemize}

Offline distillation is the simplest form of training knowledge distillation. Using this method, the teacher is trained first and its weights are frozen. Then, the student is trained on the teacher's weights. The teacher model is not updated at any point during the training of the student. This approach focuses on the algorithm in which knowledge is transferred from teacher to student, rather than the architecture of the teacher model. Offline training can also be used with a wider range of model types.

Online distillation sequentially transfers knowledge from teacher to student, rather than in blocks. The teacher model is constantly updated with new data, and then the student model is updated downstream. This results in the teacher and student model being trained simultaneously. The student improves and moves with the teacher model as the teacher model learns in real-time. This type of distillation usually incorporates some sort of feedback loop where the teacher's output updates the student and the student's output updates the teacher. This enables both models to learn on the fly. The strongest advantage of online training is its ability to handle data that changes over time. If the teacher model can react to new data, it can pass insights along to the student during training.

Self-distillation address the two main issues of both online and offline distillation: the student model's accuracy is strongly affected by the choice of teacher model, and the student model usually can't reach the teacher's level of accuracy. The self-distillation approach uses the same neural network for both the teacher and student models, integrating attention~\cite{vaswani2023attentionneed} classifiers on the hidden layers of the network. These classifiers act as the teacher model using a special loss function to update the student's scores – the additional classifiers are removed after training.

This paper will utilize offline distillation.

\subsection{Tsetlin Machine} \label{subsec:tm}

\subsubsection{Overview}
The Tsetlin Machine (TM), based on the Tsetlin Automata (TA), is a relatively new machine learning algorithm introduced by Granmo et al. in 2018~\cite{granmo2018tsetlin}. Tsetlin Machines are capable of finding sophisticated patterns in data using simple propositional logic, instead of using complex math as in neural networks. They have demonstrated applications in both classification and regression.
Tsetlin Machines enjoy efficient learning speed and low memory usage while achieving predictive performance competitive to other machine learning architectures in several benchmark datasets~\cite{abeyrathna2021massively, lei2020arithmetic}. A Tsetlin Machine is based on the multi-armed bandit problem~\cite{robbins1952some, gittins1979bandit} and the Finite State Machine~\cite{narendra2012learning}. 

Using propositional logic in Tsetlin Machines has many unique advantages over traditional machine learning algorithms. Propositional logic based learning is more interpretable and explainable, making it easier for humans to understand. Because of this, Tsetlin Machines have numerous important applications in explainable AI (xAI), a field where transparency and interpretability are of chief importance~\cite{xu2019explainable,dovsilovic2018explainable}. Tsetlin Machines can also achieve a much faster execution time over traditional machine learning algorithms that require advanced vector multiplication. Propositional logic based learning algorithms have been utilized by traditional machine learning algorithms~\cite{anjum2022localization, anjumlocalization}.

\begin{figure}
    \centering
    \includegraphics[width=0.65\linewidth]{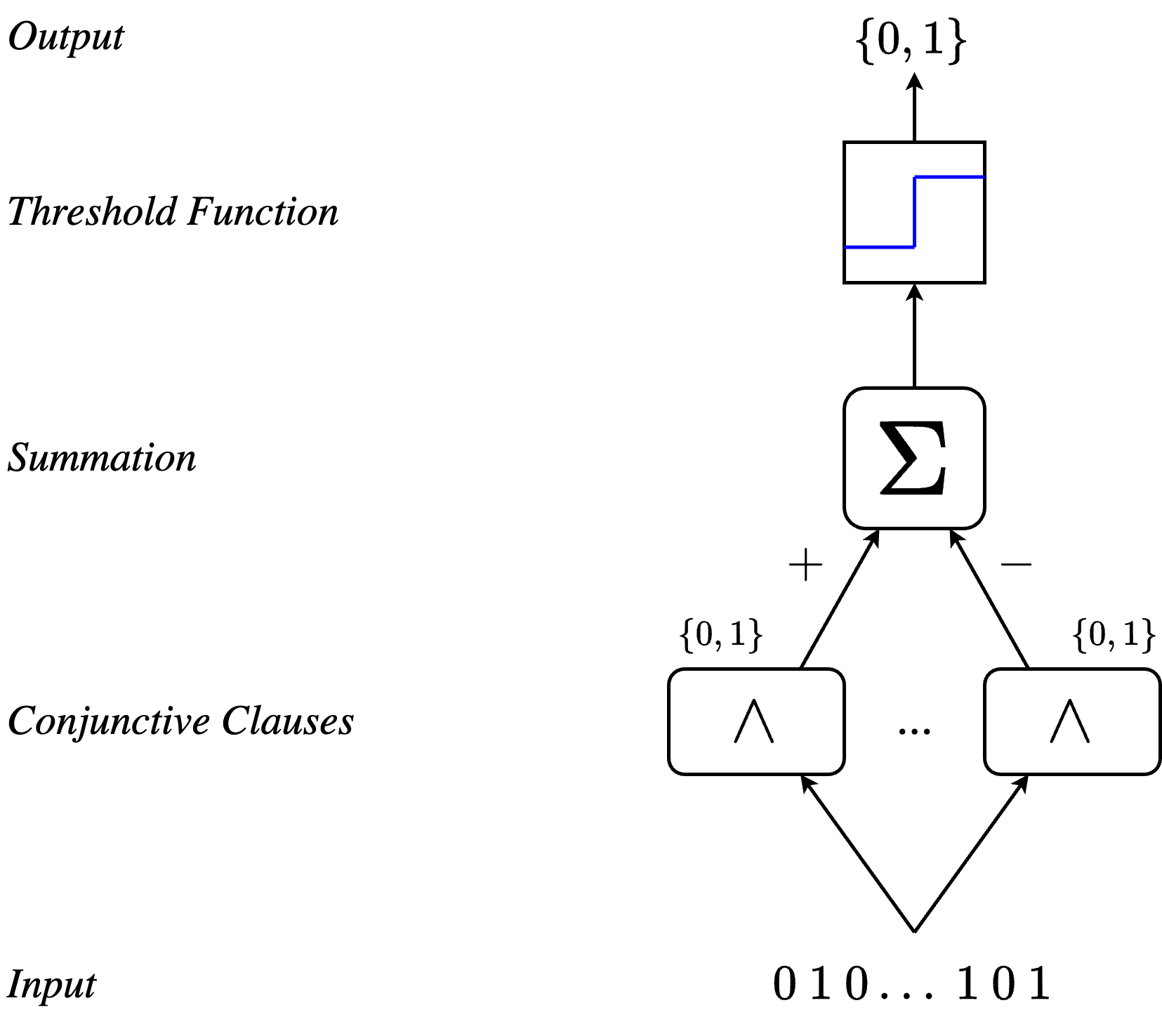}
    \caption{Inference structure for a binary output Testlin machine.}
    \label{fig:inference-structure}
\end{figure}

In Tsetlin Machines, propositional logic is used to build clauses ($C$) of literals ($L$) created using features of the input data. For example, in an image, a pixel could be a literal. The literals of the clauses are learned using feedback rules which change the literals' locations in memory during training. It is important to mention that, while a pixel may be a literal, the set of all pixels is not equal to the set of all literals. Literals are created for each feature and each negated feature, making the number of literals equal to twice the number of features in a data sample.
Clauses are created as conjunctions of memorized literals, where the predicted class of an input sample is determined by the majority sum of each clause over each class. The clause sum is calculated by a step function parameterized by threshold $T$, which sets the value where a clause becomes true or false (Figure~\ref{fig:inference-structure})~\cite{granmo2018tsetlin}. 

\subsubsection{Mathematical Definition}

Using propositional logic statements, the Tsetlin Machine is capable of solving complex pattern recognition problems~\cite{granmo2018tsetlin}. As mentioned in the previous section, a standard Tsetlin Machine is composed of an assembly of Tsetlin Automatons (TAs)~\cite{tsetlin1961behaviour}. Each TA exists as a part of a conjunctive clause of the Tsetlin Machine. In this section, we give a technical definition of the Tsetlin Machine to paint a proper background for the algorithms proposed in this paper.

First, the Tsetlin Machine takes a vector $X$ of $x\in\{0,1\}^n$ propositional variables and is defined~as:
\begin{align}
    X = (x_1, x_2, \dots, x_n) \in \{0,1\}^n
\end{align}

For example, if $n=2$ (2 features), then the components of $X$ are $x_1$ and $x_2$, and the domain space is $\{(0, 0), (0, 1),$ $(1, 0), (1, 1)\}$. 

The Tsetlin Machine is able to deduce patterns in the input data using conjunctive clauses consisting of literals in the dataset. A literal can be one of the boolean features in the input data or the negation of a feature in the input data. This can be represented as $L = l_k \in \{0, 1\}^{2n}$ where each literal $l_k$ is defined as:
\begin{align}
    l_k = \begin{cases}
    x_k & if \quad 1 \le k \le n, \\
    \neg x_{k-n} & if \quad n+1 \le k \le 2n
    \end{cases}
\end{align}

This defines the first $n$ literals as the actual variables and the next $n$ variables as the negated variables. Therefore the number of literals in a model can be simply calculated as:
\begin{equation}\label{eq:num-literals}
    |L|=2n
\end{equation}
Using the aforementioned example of $n=2$, then $L = \{x_1, x_2, \neg x_1, \neg x_2\}$.

In general, the number of clauses is a user-defined parameter and can be considered a hyperparameter of the Tsetlin Machine. This has a significant effect on accuracy~\cite{9923795}. Let the set of clauses belonging to a Tsetlin Machine be defined as $C$ and each clause be denoted as $C_j$ where $j \in \{1,...,|C|\}$. Thus, it follows that the clause is a conjunction of a subset of literals $L_j \subseteq L$:

\begin{align}
    C_j = \bigwedge_{l_k \in L_j} l_k = \prod_{l_k \in L_j} l_k
\end{align}

A clause $C_j$ evaluates to $C_j \in \{0, 1\}$ depending on the truth values of each literal that belongs to it. For example, let $|C| = 2$ with the user defined clauses being $C_1 = x_1 \wedge \neg x_2$ and $C_2 = x_1 \wedge x_2$. Then, extending what we defined previously, the literals have the index $L_1 = \{1, 4\}$ and $L_2 = \{1, 2\}$ in the full set of literals with length $|L|=2n$.

Each clause in $C$ is assigned a polarity $p \in \{-1, +1\}$, denoting whether the clause is negative or positive. Specifically, positively polarized clauses, defined as $C_j^+ \in \{0, 1\}^{\frac{|C|}{2}}$, are assigned to class $y = 1$. Negatively polarized clauses, defined as $C_j^-\in \{0, 1\}^{\frac{|C|}{2}}$, are assigned to class $y = 0$. In a typical implementation, half of the clauses are given positive polarity and the remaining half given negative polarity. It follows that the positively polarized clauses vote to classify the input sample as the target class, and inversely for the negatively polarized classes.

In order to compute the final classification decision, the clause outputs are summed:

\begin{align} \label{eq:basic_classification}
    s(X) & = \sum_{j=1}^{|C|/2}C_j^+ - \sum_{j=1}^{|C|/2}C_j^- \\
    \hat{y} & = u(s(X))  \nonumber
\end{align}
where $u(v)$ is the unit step function defined by:
\begin{equation}
    u(v) = \begin{cases}
    1 & if \quad v \ge 0 \\
    0 & otherwise
    \end{cases}
\end{equation}
The output of the summation decides the selected class $\hat{y} \in \{0, 1\}$.

\begin{figure}[t]
    \includegraphics[width=\textwidth]{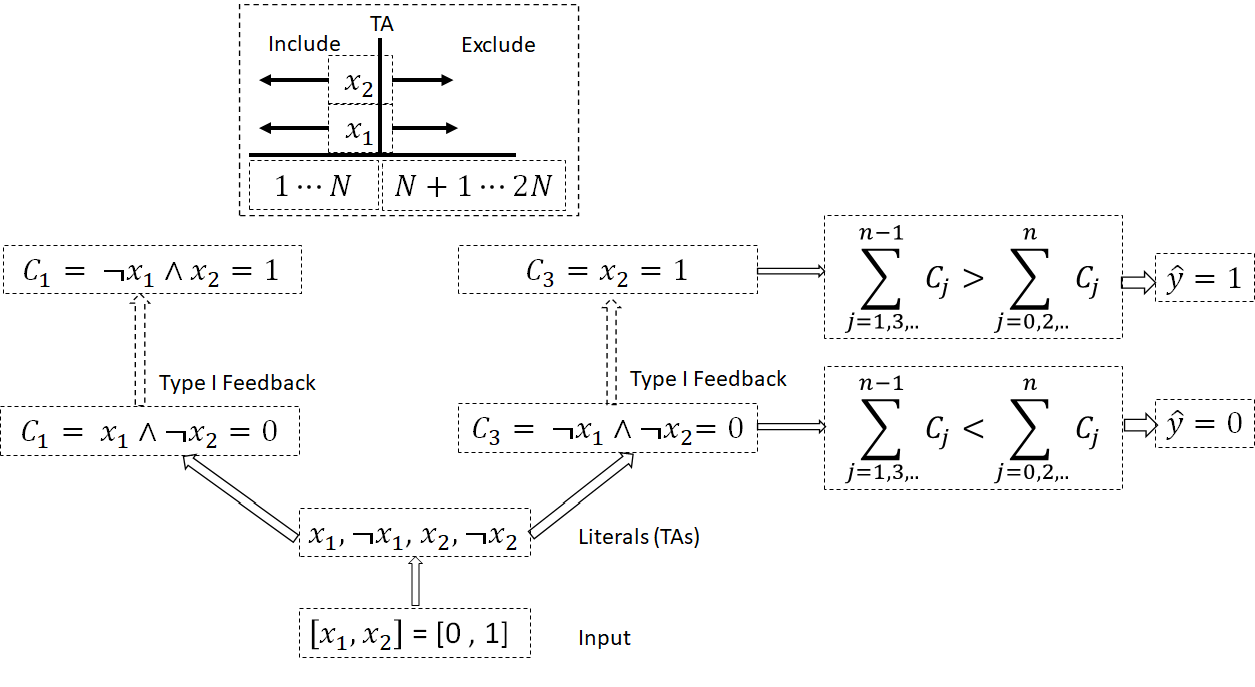}
    \caption{TM learning dynamics with input ($x_1=0, x_2=1$) and output $y=1$.}
    \label{fig:learn-dyn}
\end{figure}

For our previously defined example, Figure~\ref{fig:learn-dyn} shows two positive polarity clauses, $C_1=x_1 \wedge \neg x_2$ and $C_3=\neg x_1 \wedge \neg x_2$ (clauses with negative polarity are omitted). Both clauses evaluate to zero, resolving to a final classification of $\hat{y}=1$.

For a single training step in a Tsetlin Machine, clause sums are used instead of a typical loss function. Each clause is given feedback with a probability $\hat{p}$, determined by:
\begin{align} \label{eq:feedback-equation}
    \hat{p} = \begin{cases}
        \displaystyle\frac{T + \text{clamp}(s(X), -T, T)}{2T} & if \quad y=0 \\
        \displaystyle\frac{T - \text{clamp}(s(X), -T, T)}{2T} & if \quad y=1
    \end{cases}
\end{align}

In Equation~\eqref{eq:feedback-equation}, $s(X)$ is the sum of the clauses as defined in Equation~\eqref{eq:basic_classification}. The threshold, $T$, is a user-defined parameter. The~$\text{clamp}$ function scales the $s(X)$ between $-T$ and $T$, clipping off any overflow. This random clause selection is paramount to the Tsetlin Machine. It directs the clauses to form a distribution across the quantifiable patterns in the data in an efficient manner. Because of the $\text{clamp}$ function, the truth value of each clause is proportional to $\pm T$. Effectively, the feedback amount depends on $T$, generating more feedback when $s(X)$ is far from $T$.

\begin{figure}[t]
    \includegraphics[width=\textwidth]{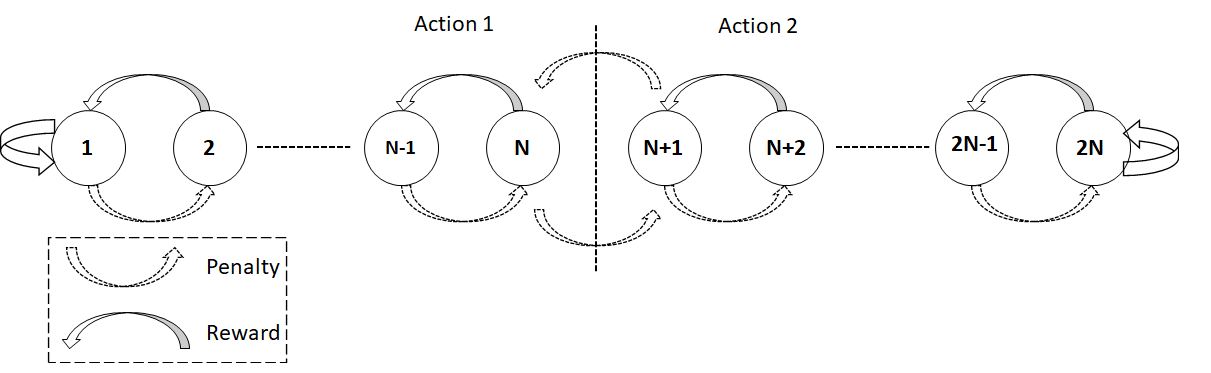}
    \caption{A two-action Tsetlin Automaton (TA) with $2N$ states.}
    \label{fig:basic-ta}
\end{figure}

A basic Tsetlin Automaton is visualized in the upper boxed area of Figure~\ref{fig:learn-dyn}. The TA controls which actions are taken during learning steps, deciding whether to include or exclude certain literals in a clause based on the feedback received. As described previously, feedback can be classified as either Type I or Type II feedback. Figure~\ref{fig:basic-ta} shows a more detailed Tsetlin Automaton with $2N$ states. When the TA is in any state $1 \dots N$, the TA excludes the literal (Action 1). Conversely, in states $N+1 \dots 2N$, the TA includes the literal (Action 2). Thus, when a TA receives Type I feedback, the literal represented by the TA moves further into being included; when a TA receives Type II feedback, the literal moves further into being excluded.

Type I feedback works to reduce the frequency of overfitting and false negatives by recognizing frequent patterns. Type II feedback reduces the occurrence of false positives by enhancing the discrimination between samples belonging to different classes.

During Type I feedback, positively polarized clauses receive positive feedback when $y=1$ and negative feedback on negatively polarized clauses when $y=0$. Type I feedback has two subtypes: Type IA (include) feedback increments the literal's position in memory, and Type IB (exclude) feedback decrements the literal's position in~memory. 

Type I feedback is parametrized by specificity $s$ where $s \ge 1$. Define the TA state action on the $k^{th}$ literal and the $j^{th}$ clause by $a_{j,k}$. State $a_{j,k}$ receives Type IA feedback with probability $\frac{s-1}{s}$ and Type IB feedback with probability $\frac{1}{s}$. When the TA receives Type IA feedback, it increments by 1, and inversely for Type IB feedback. The probability design pushes Type IA and Type IB feedback to include and exclude literals, respectively.

Type II feedback is applied to clauses with positive polarity when $y=0$ and to clauses with negative polarity when $y=1$. TA states are not updated when the clause output is 0. Conversely, when the clause output is 1, all included literals with value 0 in state $a_{j,k}$ are incremented by 1. Type II feedback is focused on literals that differentiate the clause output between $y=0$ and $y=1$.

\subsubsection{Modifications}
\paragraph{Weighted Tsetlin Machine.}
The standard TM was extended with the Weighted Tsetlin Machine (WTM), introduced in~\cite{phoulady2019weighted}. This modified the original summation by multiplying a weight with each clause like shown:
\begin{align} \label{eq:weighted-classification}
    s(X) = \sum_{j=1}^{|C|/2} w_j^+C_j^+ - \sum_{j=1}^{|C|/2}w_j^-C_j^-
\end{align}
All weights are initialized at 1.0
\begin{align}
    w_j^+&\gets1.0,\\
    w_j^-&\gets1.0.
\end{align}
Therefore, at the start of training, clause behavior is identical to that of a standard TM. 
Weight updates are controlled by Type I and Type II feedback with a learning rate $\gamma\in[0,\infty]$. Using Type I feedback, each clause returning 1 has its respective weight multiplied by $1+\gamma$:
\begin{align}
    w_j^+ &\gets w_j^+ \cdot (1 + \gamma), \quad \text{if } C_j^+(\mathbf{x}) = 1,\\
    w_j^- &\gets w_j^- \cdot (1 + \gamma), \quad \text{if } C_j^-(\mathbf{x}) = 1.
\end{align}
Likewise, using Type II feedback, the weights are instead divided by $1+\gamma$:
\begin{align}
    w_j^+ &\gets w_j^+ \cdot (1 + \gamma), \quad \text{if } C_j^+(\mathbf{x}) = 1,\\
    w_j^- &\gets w_j^- \cdot (1 + \gamma), \quad \text{if } C_j^-(\mathbf{x}) = 1.
\end{align}
Clauses that evaluate to 0 do not have their weights updated:
\begin{align}
    w_j^+ &\gets w_j^+, \quad \text{if } C_j^+(\mathbf{x}) = 0,\\
    w_j^- &\gets w_j^-, \quad \text{if } C_j^-(\mathbf{x}) = 0.
\end{align}
These weights help quantify and reinforce true positive frequent patterns while diminishing the impact of false positive patterns on the classification decision. 

\paragraph{Multi-Class Tsetlin Machine.}
So far, the Tsetlin Machine has only been described as a binary classifier. However, using a Multi-Class Tsetlin Machine (MCTM) architecture, a Tsetlin Machine can learn to distinguish between any number of classes. This is done by using a collection of Tsetlin automaton teams, one for each class. Instead of using a step function to discriminate between classes 0 and 1 as shown in Equation \eqref{eq:basic_classification}, an $\mathrm{argmax}$ function is used to determine the index of the clause sum with the largest value. 

Let $\zeta$ be the number of classes in a dataset. Then \eqref{eq:basic_classification} is modified to be

\begin{equation} \label{eq:multi_classification}
   \hat{y}  = \underset{i=1,\dots,\zeta}{\mathrm{argmax}} \left( \sum_{j=1}^{|C|/2}C_j^+ - \sum_{j=1}^{|C|/2}C_j^- \right)
\end{equation}

Training is done in the same way as the vanilla Tsetlin Machine, aside from a crucial modification. Assume that $y=i$ for the current sample $(X,y)$. The Tsetlin Automata teams belonging to class $i$ are trained in the same way as per $y=1$ described in Equation \eqref{eq:feedback-equation}. However, a second, randomly selected class $q\neq i$ is chosen. The Tsetlin Automata teams belonging to class $q$ are then trained using $y=0$ as described for the vanilla Tsetlin Machine.

\paragraph{Weighted Multi-Class Tsetlin Machine.} Note that weighted Tsetlin Machines and multi-class Tsetlin Machines are not mutually exclusive. We will be using  weighted, multi-class Tsetlin Machines for all experiments in this paper. The following equation shows how classification is determined in a weighted, multi-class Tsetlin Machine (WMCTM) with $\zeta$ classes.
\begin{align} \label{eq:weighted-multi-classification}
   \hat{y} & = \underset{i=1,\dots,\zeta}{\mathrm{argmax}} \left( \sum_{j=1}^{|C|/2} w_j^+C_j^+ - \sum_{j=1}^{|C|/2}w_j^-C_j^-\right)
\end{align}
\clearpage
\section{Problem Statement}

Although novel Tsetlin Machines have made great strides in classification tasks, there are still many paths to optimization. The accuracy of a Tsetlin Machine increases with the size of its parameters, making accuracy gains more challenging in scenarios where memory and processing power are restricted. The size of a model is also inversely correlated with its training and inference speed. This causes problems in areas such as edge computing or FPGA development, where accuracy gains may not be feasible due to latency and hardware requirements~\cite{zhang2021low}.

Existing research in the field of knowledge distillation has proven that a large neural network can have its most important weights distilled down to a smaller model with a nonlinear drop in accuracy. This poses the question: can a variation of traditional knowledge distillation methods be successfully applied to Tsetlin Machines? 

\section{Research Objectives}

The principal objective of this research is to develop a novel approach to knowledge distillation in Tsetlin Machines, a subject in which no research paper currently explores (at date of writing). We seek to create a novel feature- and response-based knowledge distillation method and evaluate its effectiveness. We will compare two novel approaches to knowledge distillation and evaluate their effectiveness. As different datasets are optimized for different model parameters, we will measure our methods over multiple datasets and compare their strengths and weaknesses. We will score our method based on accuracy and execution time, compared between different baseline models and visualized in tables and charts. In addition, we will generate activation maps for each Tsetlin Machine to illustrate the knowledge transfer from the teacher to the student.

\chapter{Literature Review}

\section{Knowledge Distillation}

Knowledge distillation (KD) was initially introduced in 2015 as a way to improve model generalization without requiring a large number of supporting models~\cite{hinton2015distillingknowledgeneuralnetwork}. This paper serves as the starting point for future knowledge distillation papers.

A key approach was introduced by Gao et al. to first transfer the backbone knowledge from the teacher to the student, effectively splitting the training process into two parts~\cite{gao2019embarrassinglysimpleapproachknowledge}. This method significantly narrows the accuracy gap between teacher and student.

Knowledge distillation was supplemented by attention with Channel Distillation (CD) and Guided Knowledge Distillation (GKD)~\cite{channeldistill, DBLP:journals/corr/VaswaniSPUJGKP17}. GKD filters teacher's output to only impart correct results on the student. This method also introduced Early Teacher Decay (ETD), which gradually reduces the weight of the teacher's influence during training.

In the natural language processing (NLP) domain, knowledge distillation was implemented using an unlabeled transfer set constructed from diverse, pre-trained language models~\cite{tang-etal-2019-natural}. This approach helped the student model generalize more effectively, even beating OpenAI's GPT while using a bidirectional LSTM~\cite{radford2019language, cui2019deepbidirectionalunidirectionallstm}. 

Knowledge distillation has also been used in ranking systems. RankDistil is a distillation method for top-k ranking, using statistical methods combined with knowledge distillation to optimize ranking systems~\cite{pmlr-v130-reddi21a}. This method is specifically targeted at ranking problems with a large number of items to rank.

Another approach, Born Again Neural Networks (BANs), studies knowledge distillation from a fresh perspective~\cite{furlanello2018bornneuralnetworks}. BANs feature a student and teacher model with identical parameters. Using KD in this fashion boosts accuracy from shared information. 

One unique method in KD is called self-distillation~\cite{zhang2019teacherimproveperformanceconvolutional}. Self-distillation progressively shrinks the size of a network, improving performance and accuracy within the same model.

Smaller models can also be initialized from the pretrained weights of the teacher~\cite{xu2023initializingmodelslargerones}. This efficient method can reduce training time and improve accuracy in the student.

\section{Tsetlin Machine}

In 2018, Granmo et al. introduced the novel Tsetlin Machine (TM) as a successful approach to the multi-armed bandit problem~\cite{granmo2018tsetlin, bouneffouf2023multi, bouneffouf2020survey}. Since its introduction, there have been numerous significant works improving on the base idea.

The Tsetlin Machine was reintroduced as a contextual bandit algorithm, with two learning algorithms proposed: Thompson sampling~\cite{seraj2022tsetlin} and $\epsilon$-greedy arm selection. 

The vanilla Tsetlin Machine was extended into the image classification domain with the convolutional Tsetlin Machine~\cite{granmo2019convolutional, tunheim2023convolutional}. The convolutional Tsetlin Machine utilized clauses as a convolution filter over each image, incorporating a key technique from neural network image classification~\cite{convolutional2015introduction}.

Tsetlin Machines have also been extended to regression-based problems~\cite{darshana2020regression}. Regression-based Tsetlin Machines are also capable of outputting continuous data instead of categorical data. This type of Tsetlin Machine utilizes a special voting method to decompose complex data patterns.

Rather than being restricted to boolean-output problems, a multi-output coalesced Tsetlin Machine was proposed that can learn both the weights and composition of each clause~\cite{glimsdal2021coalesced}. This type of Tsetlin Machine employs Stochastic Searching on the Line (SSL)~\cite{oommen1997stochastic}.

Tsetlin Machines have also been shown to have great potential in the natural language processing (NLP) domain. This includes text encoding~\cite{bhattarai2023tsetlin}, text classification~\cite{saha2022relational, saha2023using, berge2019using}, sentiment analysis~\cite{yadav2021sentiment}, question classification~\cite{nicolae2021question}, and fake news detection~\cite{bhattarai2022fakenews}.

 The healthcare domain is another area that Tsetlin Machines have been applied to. A Tsetlin Machine architecture was implemented for premature ventricular contraction identification by analyzing long-term ECG signals~\cite{zhang2023interpretable}. Using a regression-based TM was proposed for predicting disease outbreaks~\cite{darshana2020regression}. Intelligent prefiltering over measurements from healthcare devices was also explored as a potential TM application, using Principal Component Analysis and Partial Least Squares Regression to improve performance~\cite{jenul2022component}.

A label-critical Tsetlin Machine was proposed that employed twin label-critical Tsetlin Automata (TA)s. The label-critic TA used a logical self-corrected TM clause as a guide to the correct label for each sample~\cite{abouzeid2022label}.

One method that has improved the feasibility of Tsetlin Machines is incorporating a drop clause probability~\cite{sharma2023drop}. A TM with a drop clause probability prunes clauses randomly with a rate of a specified probability. This can improve the robustness, accuracy, and latency of a Tsetlin Machine, quite similar to a dropout layer in a traditional neural network~\cite{JMLR:dropout}. It is emphasized that these clauses are dropped completely randomly.

One significant improvement to the vanilla TM was the addition of weighted clauses. The Weighted Tsetlin Machine (WTM) pairs a weight with each clause to give more consideration to the most useful clauses~\cite{phoulady2019weighted}. This approach also reduces computation time and memory usage.

In order to discriminate between closely similar clauses, focused negative sampling (FNS) was introduced~\cite{glimsdal2022focused}. This method reduces training time.

A new learning method, Type III feedback, was proposed in 2019~\cite{granmo2023learning}. This approach is a more efficient way of pruning clauses using the Markov boundary, and supplements the existing Type I and Type II feedback already being used in Tsetlin Machines.

Using a regularizer in the TM has been found to improve its performance and generalization. This method is called Regularized TM (RegTM), and supports both a moving average regularizer and a weighted average regularizer~\cite{anjum2024novel}. The authors also proposed the use of a softmax (sigmoid) function for classification as an alternative to the clause sum~\cite{softmax}.

Another significant work explored the effects of a multi-layer Tsetlin Machine architecture~\cite{gorbenko2024multi}. This approach showed that accuracy can be increased by adopting a hierarchical feature learning approach.

One proposed method to accelerate Tsetlin Machine computation involves the inclusion of Absorbing Automata~\cite{bhattarai2023contractingtsetlinmachineabsorbing}. This method contracts the TM by making the learning scheme absorbing rather than ergodic. Clauses become set in their outputs as training goes on, requiring fewer and fewer clauses to be computed per epoch.

A variant on the TM called the Clause Size Constrained TM (CSC-TM) was also introduced~\cite{abeyrathna2023building}. The CSC-TM focuses on improving clause efficiency by placing a soft constraint on the clause size, restricting the amount of literals that can be gathered. Another work explored the optimal clause count through a runtime pruning system~\cite{9923795}.

\chapter{Methodology}

In this section, we explore two different ways that knowledge distillation can be implemented in a Tsetlin Machine. We introduce a naive method based on the Multi-layer Tsetlin Machine described in~\cite{gorbenko2024multi} and a more involved version based on comparing probability distributions. The former method is called Clause-Based Knowledge Distillation (CKD), and the latter is called Distribution-Enhanced Knowledge Distillation (DKD). CKD is a feature-based method that uses a multi-layer approach to reduce the size at each layer to reduce the training time. DKD is more similar to knowledge distillation in neural networks, utilizing the probability distributions from the teacher model for each sample to supplement the input dataset.

While developing distribution-enhanced knowledge distillation, we started by experimenting with the described CKD method. We then explored the more involved DKD approach. CKD was easily implemented using standard available libraries, which aided our rapid development early on. As we describe later on, we believe DKD has more use cases in the Tsetlin Machine domain.

Comparing both methods illustrates our development process and aids our selection of an optimal approach to implementing knowledge distillation in Tsetlin Machines. 

We will use $TM_T$ to represent the teacher model, $TM_S$ to represent the student model baseline, and $TM_D$ to represent the distilled model. $TM_D$ and $TM_S$ are parametrically identical --- $TM_S$ only exists to compare the accuracy of a model \textit{without} distillation to the identically-sized model $TM_D$ \textit{with} distillation.

\section{Clause-Based Knowledge Distillation}\label{sec:ckd-sec}

As previously discussed, clause-based knowledge distillation is based on the hierarchical architecture proposed in \cite{gorbenko2024multi}. However, unlike the proposed approach that aims to add multiple layers to TM architecture, our method focuses more on its applications to knowledge distillation in the TM domain, where information is transferred from the teacher model to the student model. As a result, the student model performance becomes comparable in accuracy to the teacher model while maintaining the training time of the student model.

In Tsetlin Machines, feedback is based on the expected class and the generated clause outputs. As a result, knowledge distillation in TM can be represented in the transfer of clauses from the teacher model $TM_T$ to the student model $TM_S$. Previous research has shown that the transferring knowledge from the teacher to the student model can result in incremental accuracy gains for the student model, even when the teacher model is trained for only a few epochs \cite{lei2020arithmetic}. Hence, a knowledge distillation model in a Tsetlin Machine would consist of training the teacher for a few epochs $E_T$, and then using the clause outputs generated from the teacher model as input data to train the student model. 

As explained in Section~\ref{subsec:tm}, each class's clauses decisions are independent from other clauses. By using a second distilled $TM_D$ that takes the output of the first teacher $TM_T$, we can obtain an accuracy boost from the enhanced context of the independent clauses in the teacher, even when the distilled TM has significantly fewer clauses than the teacher. Since the clause outputs in each class of the teacher are independent of each other, feeding those clause outputs into a second TM as literals gives the model a broader picture of the data. This can potentially reduce training time and increase accuracy. 

\subsection{Clause Output Generation}\label{sec:clause-output}

\begin{figure}[t]
    \centering
    \includegraphics[width=0.8\textwidth]{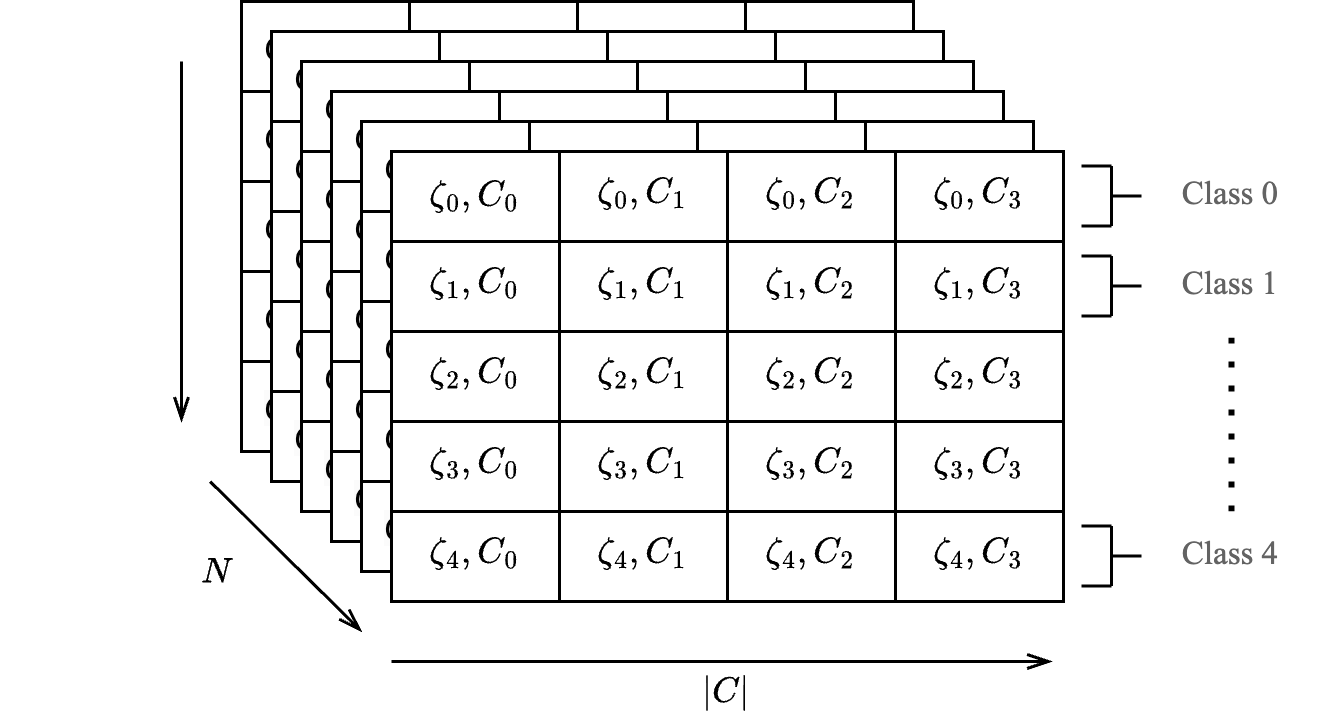}
    \caption{Layout in memory of the clause outputted by the teacher.}
    \label{fig:teacher-clause-output}
\end{figure}

Each TM clause of the teacher model can be represented as a conjunction of $(x_k\vee\neg TA_k^{T})$ and $(\neg x_k\vee\neg TA_{2n+k}^{T})$ disjunctions:

\begin{equation}\label{eq:clause-setup}
    Clause_{\textit{Class}=0.9, j=0, C-1}^{\textbf{T}} =
    \bigwedge_{k=0}^{n-1} (x_k \vee \neg TA_{k}^{T}) \wedge (\neg x_k \vee \neg TA_{2n+k}^{T})
\end{equation}

where $x_k$ is a propositional variable (for images, a single pixel of a binarized black and white image) and $n$ is the number of features (pixels in this example). Recall from Equation~\eqref{eq:num-literals} that the number of literals can be calculated as twice the number of features, or $|L|=2n$. The output from the teacher model becomes the input to the student model. The teacher model transforms the input such that for each class there is a set of classes. For example, a 2D input shape of $(N,n)$ converts to a shape of $(N, \zeta \times |C|)$ where $N$ is the number of samples, $\zeta$ is the set of classes and $|C|$ is the number of clauses in the teacher model. Figure~\ref{fig:teacher-clause-output} illustrates this output. 

This essentially creates a new binary feature for each clause in each class, where the value of each new feature is determined by whether the corresponding clause evaluates to true or false for the input example. This clause output feature space is valuable because it represents the input data in terms of the learned patterns (clauses) that the Tsetlin~Machine has identified as important for classification.

In simple terms, the teacher model is increasing the input data size of the distilled model from $n_S$ to
\begin{equation}\label{eq:literals-d}
    n_{D}=\zeta*|C_T|
\end{equation}
This will result in an increase to the accuracy of the student model at the cost of increased training time and memory requirements if $n_D>n_S$. This is contrary to traditional knowledge distillation, where the objective is to not only improve accuracy of the student model, but to also reduce the memory and time requirements. When the difference between the number of teacher clauses $|C_T|$ and student clauses $|C_S|$ is large, the increase in training time is even larger, sometimes even surpassing the training time of the teacher model.

We can also present this in terms of the information theory perspective. The information in a standard Tsetlin Machine can be assumed to be based on the number of literals and the clauses. As the number of literals and clauses is increased, so does the measure of information contained. We can express the measure of the information in a Tsetlin Machine as:

\begin{equation} \label{eq:general-info}
   I=\frac{1}{|L|*|C|}log(\frac{1}{|L|*|C|})
\end{equation}

 where $|L|$ is the number of literals and $|C|$ is the number of clauses.

If the difference between the number of teacher clauses $C_T$ and student clauses $C_S$ is large, it follows that the training time difference between the teacher and student model will also be large. However, as the number of classes grows, the training time difference will grow $\zeta$-fold. If the distilled model takes more time than the teacher model for less accuracy, it doesn't make sense to use knowledge distillation.

To combat this, we propose Probabilistic Clause Downsampling (PCD), detailed in Algorithm \ref{algo:pcd}. 

\subsection{Probabilistic Clause Downsampling}

\begin{figure}[t]
    \centering
    \includegraphics[width=0.75\textwidth]{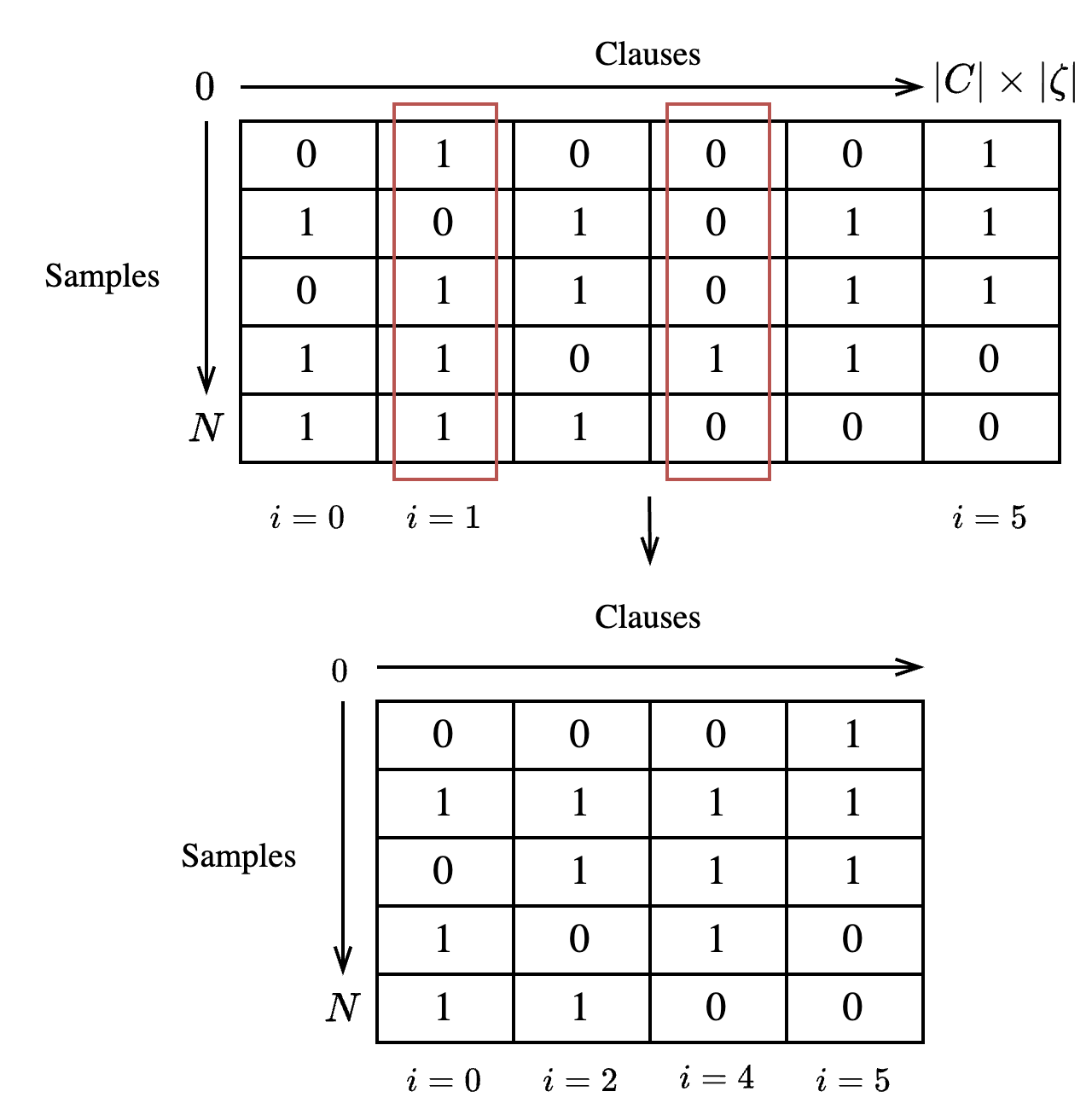}
    \caption{Probabilistic clause downsampling (PCD) in action on a small matrix.}
    \label{fig:downsampling-in-action}
\end{figure}

Probabilistic Clause Downsampling aims to find repeated information in the clause outputs of the teacher model at the Markov boundary~\cite{granmo2023learning}. Recall that the output of the teacher model is of shape $(N, \zeta \times |C_T|)$, so $X_\text{train}$ for $TM_D$ is in space $\mathbb{R}^{N\times(\zeta\times|C_T|)}$. Given that we now have clause outputs for each class, there is bound to be some redundant information contained in the output. We can find this information in a manner similar to item set mining \cite{itemset}. Each clause teacher's model output is aligned over each sample (Figure \ref{fig:downsampling-in-action}). 
Next, all columns that contain the same value with at least $1-\delta$ probability are removed. For example, in Figure~\ref{fig:downsampling-in-action}, when $\delta=0.2$, columns 1 and 4 are removed. We can verify this through Equation~\eqref{eq:general-info}. An optimal downsampling rate $\delta$ would have student model training time less than that of the teacher model while keeping its accuracy as close to the teacher model as possible. Note that generating the columns to drop from $\delta$ is best performed over the training set and then applied to the test set (as shown in the algorithm); it is a trainable method. Keeping with current naming convention, we will refer to a distilled Tsetlin Machine using PCD as $TM_{PCD}$.

\begin{figure}[t]
    \centering
    \includegraphics[width=0.9\textwidth]{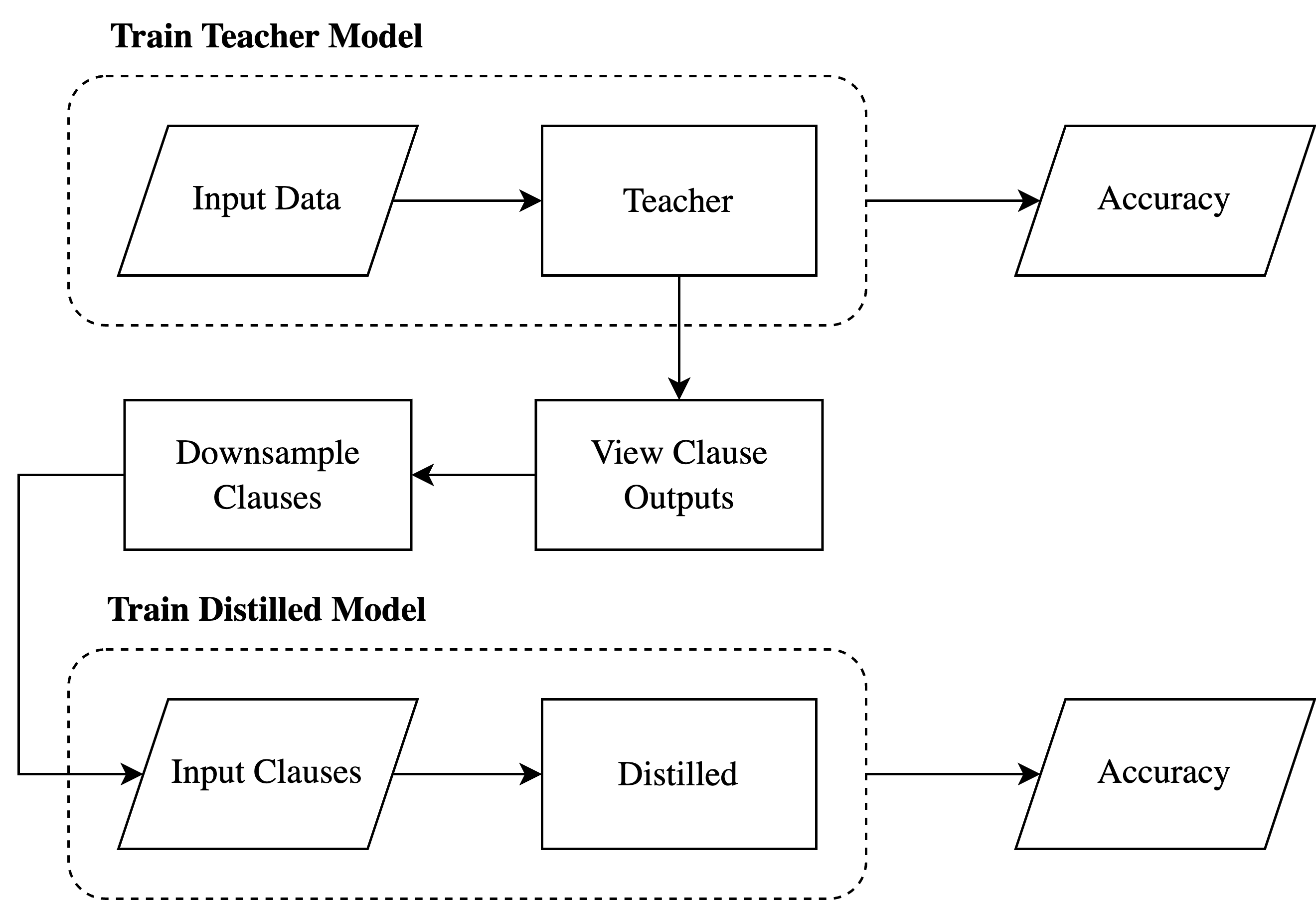}
    \caption{Data flow in clause-based knowledge distillation.}
    \label{fig:distillation-flow-chart}
\end{figure}

Figure~\ref{fig:distillation-flow-chart} shows the typical flow of data through a system using clause-based knowledge distillation with PCD.

\begin{algorithm}
\caption{Probabilistic Clause Downsampling.}
\label{algo:pcd} 
\begin{algorithmic}[1]
\Require $X_\text{train} \in \mathbb{R}^{n \times m}$: Training data transformed by teacher TM's clauses. 
\Require $X_\text{test} \in \mathbb{R}^{k \times m}$: Test data transformed by teacher TM's clauses
\Require $\delta \in [0,1)$: Threshold parameter for clause pruning
\Ensure Returns downsampled training and test data matrices
\Function{PCD}{$X_\text{train}, X_\text{test}, \delta$}

\State $\mathbf{s} \gets \sum_{i=1}^{n} X_\text{train}[i,:]$ \Comment{$\mathbf{s} \in \mathbb{R}^m$: Sum of activations per clause}
\State $\mathbf{p} \gets \mathbf{s} / n$ \Comment{$\mathbf{p} \in [0,1]^m$: Normalized activation probabilities}

\State $C^\text{hi} \gets \{j \in \{1,2,\ldots,m\} : p_j > (1-\delta)\}$ \Comment{Set of highly active clauses}
\State $C^\text{lo} \gets \{j \in \{1,2,\ldots,m\} : p_j < \delta\}$ \Comment{Set of rarely active clauses}
\State $C^{\text{drop}} \gets C^\text{hi} \cup C^\text{lo}$ \Comment{Set of clauses to eliminate}

\State $\mathcal{J} \gets \{1,2,\ldots,m\} \setminus C^{\text{drop}}$ \Comment{Set of indices of columns to retain}
\State $X'_\text{train} \gets X_\text{train}[:, \mathcal{J}] \in \mathbb{R}^{n \times |\mathcal{J}|}$ \Comment{Project to reduced feature space}
\State $X'_\text{test} \gets X_\text{test}[:, \mathcal{J}] \in \mathbb{R}^{k \times |\mathcal{J}|}$ \Comment{Apply same projection}

\State \Return $(X'_\text{train}, X'_\text{test})$
\EndFunction
\end{algorithmic}
\end{algorithm}

Although PCD does improve the performance of this approach, CKD still has one very significant limitation: the teacher TM must still be retained even after the distilled TM is trained. Since the input to the distilled TM is the based on the output of the teacher, any new data must first pass through the teacher TM to get the clause outputs. Therefore, this distilled TM in this approach can never be faster than the teacher during inference. However, this approach could still potentially be useful in applications where training time is critical.

\section{Distribution-Enhanced Knowledge Distillation}

Knowledge distillation can also be implemented in Tsetlin Machines by using a similar approach to knowledge distillation in neural networks, along with some added modifications to fit with Tsetlin Machines. Traditional neural networks use the logits obtained after the sigmoid (softmax) function of the teacher model (response-based) or from the hidden layers of the teacher model to train the student model~\cite{Gou_2021}. These logits and hidden layers gain their information through a loss function with backpropagation. However, as there is no such backpropagation mechanism in Tsetlin Machines, this information is typically not available. We circumvent this limitation by creating probability distributions from the normalized class sums of the teacher and initializing the student with the most important clauses in the teacher. This results in a hybrid feature- and response-based implementation that significantly increases the accuracy of the student with little to no detrimental effect on latency~\cite{anjum2024novel, xu2023initializingmodelslargerones}. 

Like traditional knowledge distillation, our novel approach is guided by the idea that the most valuable information can be gleaned from the teacher and then passed on to the student. Our implementation combines two methods: initialization of the student with the most important clauses from the teacher, and supplementation of the student's fit method with probability distributions made from the teacher's output samples. Additionally, we introduce several intuitive and tunable parameters:
\begin{enumerate}
    \item Temperature (entropy of teacher's output distributions): $\tau\in(0,\infty)$
    \item Balance (influence of teacher in training): $\alpha\in[0,1]$
    \item Weight transfer (influence of teacher's clause weights in initialization): $z\in[0,1]$
\end{enumerate}
These parameters allow the algorithm more granular control over the teacher's influence on the student. As in traditional KD  (and Section~\ref{sec:ckd-sec}), the teacher is first trained for $E_T$ epochs, and then the student is initialized and trained using supplemental information from the teacher for $E_S$ epochs.

Distribution-enhanced knowledge distillation is comparatively very similar to knowledge distillation in neural networks. This approach has the advantage of not requiring the teacher TM for inference once training is completed. The distilled TM can be completely disconnected from the teacher, allowing the distilled TM to work in environments that may not support the size of the larger teacher TM.

\subsection{Clause Initialization}

The first part of this hybrid knowledge distillation occurs before training the distilled model. We can express the teacher Tsetlin Machine that has a set of clauses $C_T $ as $TM_T\ni C_T$ and the student as $TM_S\ni C_S$, respectively. Since knowledge distillation is designed to use a larger model to improve a smaller model, we can assume $|C_T|>|C_S|$. 

As shown previously, a smaller neural network can be initialized with a subset of weights from a larger model~\cite{xu2023initializingmodelslargerones}. The only component of a Tsetlin Machine analogous to the hidden layers of a neural network is a TM's set of clauses, which can be transferred between models. Thus, it follows that a student $TM_S$ could copy up to $|C_S|$ clauses of the teacher into itself, initializing the student with pretrained information. This is considered a form of feature-based knowledge distillation.

In order to transfer these clauses, we propose IntelligentTransfer (Algorithm \ref{algo:init-clauses}) for selecting the top-$|C_S|$ clauses of $TM_T$ by their order of influence on the teacher's decisions. Let $\zeta$ be the number of output classes in a Tsetlin Machine and $z$ be the fraction of clauses to allocate by weight, with $1-z$ being the fraction of clauses to allocate by diversity. To allocate by weight, the set of clauses in the teacher $C_T$ are sorted by their clause weights. Let $\lambda_{weight}$ be the set of indices of the top-$\lfloor z*|C_S|\rfloor$ clauses that are selected from $C_T$ and $\Omega_{remain}$ be the set of clauses with indices not in $\lambda_{weight}$.

To calculate diversity, a calculation is done on each clause in each class $k$. Iterating over the indices and weights $(j,w_j)\in\Omega_{remain}$, let $a_j=\displaystyle\frac{|A^k_j|_1}{|A^k_j|_0+|A^k_j|_1}$, where $A^k$ is the set of TAs of the teacher assigned to class $k$. This calculates the ratio of included Tsetlin Automata (TAs) to the total number of TAs available per clause. $a_j$ measures the relative activity of a clause in relation to the input, with a higher activity corresponding to a clause covering a wider range of features in the data. Recall that there are two TAs for each literal in the input data. Next, let $w'_j=\displaystyle\frac{w_j}{\max(W^k)}$, where $W^k$ is the set of clause weights for class $k$ in the teacher. This normalizes the clause weight over all clause weights, putting $w'_j\in[0,1]$. Combining $a_j$ and $w'_j$, let $v= w'_j \cdot a_j$. This ensures that weight is still an equal factor in addition to the diversity score. Each combined weight and index are joined in set $D$. Next, $D$ is sorted by score in descending order. Let $\lambda_{diverse}$ be the first $(|C_S|-n_{direct})$-indices from $D$ and subsequently $\lambda_{selected}$ be the union of $\lambda_{diverse}\cup \lambda_{weight}$. Then, every clause and weight in $\lambda_{selected}$ is transferred to the student. This enables the student to learn the most influential clauses from the teacher.

Changing the weight transfer parameter $z$ will affect which clauses are chosen from the teacher. If $z=0$, diversity scoring will be factored into every transferred clause. If $z=1$, clauses will only be selected based on weight. We find an optimal value at $z=0.2$, allowing a balance from both ends of the spectrum. 

\begin{algorithm}
\caption{Knowledge Transfer from Teacher to Student Tsetlin Machine.}
\label{algo:init-clauses}
\begin{algorithmic}[1]
\Require $TM_T$: trained teacher Tsetlin Machine
\Require $TM_S$: student  Tsetlin Machine
\Require $z \in [0,1]$: portion of clauses to select by weight vs diversity
\Function{IntelligentTransfer}{$TM_T, z$}

\For{each class $k \in \{1,\ldots,\zeta\}$} \Comment{Iterate for each class}
    \LineComment{Extract teacher's learned patterns for this class}
    \State $W^k, A^k \gets$ teacher state for class $k$ \Comment{$W^k$: clause weights, $A^k$: TA states}
    \\
    \LineComment{Create initial pool containing all teacher clauses with their weights}
    \State $\Omega \gets \{(j, W^k_j) : j \in [1,|W^k|]\}$ \Comment{Pairs of (index, weight)}
    \\
    \LineComment{Phase 1: Select strongest clauses directly by weight}
    \State $n_{direct} \gets \max(1, \lfloor z \cdot |C_S| \rfloor)$ \Comment{At least one clause by weight}
    \State Sort $\Omega$ by weight in descending order
    \State $\lambda_{weight} \gets \text{first }n_{direct}\text{ indices from } \Omega$ \Comment{Best clauses by weight}
    \\
    \LineComment{Phase 2: Select remaining clauses using diversity scoring}
    \State $n_{diverse} \gets |C_S| - n_{direct}$ \Comment{How many more clauses needed}
    \State $\Omega_{remain} \gets \{(j, w) \in \Omega : j \not\in \lambda_{weight}\}$ \Comment{Exclude already selected}
    \State $D \gets \emptyset$ \Comment{Will store (index, diversity score) pairs}
    \\
    \LineComment{Calculate diversity scores for remaining clauses}
    \For{$(j, w_j) \in \Omega_{remain}$}
        \Comment{Measure clause complexity via TA activation ratio}
        \State $a_j \gets \displaystyle\frac{|A^k_j|_1}{|A^k_j|_0+|A^k_j|_1}$\Comment{Active TAs divided by total TAs}
        \State $w'_j \gets \displaystyle\frac{w_j}{\max(W^k)}$ \Comment{Normalize weight to [0,1]}
        \\
        \LineComment{Combine weight and diversity with baseline preservation}
        \State $v\gets w'_j \cdot a_j$ \Comment{Multiply both weights}
        \State $D \gets D \cup \{j, v\}$ \Comment{Append to $D$}
        
    \EndFor
    \\
    \LineComment{Select most diverse high-weight clauses}
    \State Sort $D$ by score in descending order
    \State $\lambda_{diverse} \gets \text{first } n_{diverse} \text{ indices from } D$
    \\
    \LineComment{Combine both selection methods}
    \State $\lambda_{selected} \gets \lambda_{weight} \cup \lambda_{diverse}$
    \\
    \LineComment{Transfer selected clauses to student, maintaining order}
    \For{$m \gets 1$ to $|\lambda_{selected}|$}
        \State Transfer clause and weight at $\lambda_{selected}[m]$ to student $TM_S$ position $m$
    \EndFor
\EndFor

\EndFunction
\end{algorithmic}
\end{algorithm}

\subsection{Soft Label Generation}

\begin{figure}[t]
    \centering
    \includegraphics[width=0.8\textwidth]{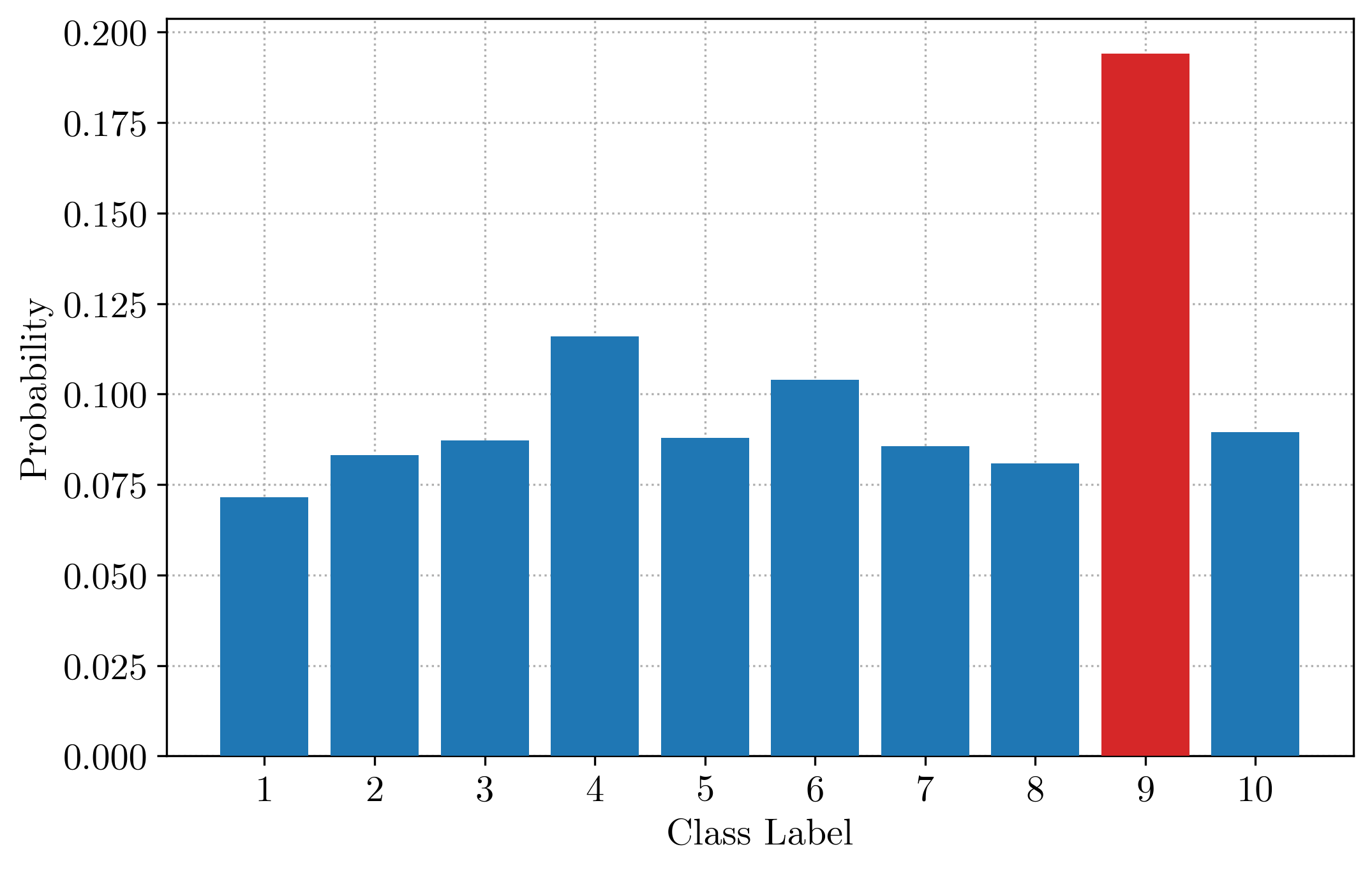}
    \caption{Visualization of the distribution defined in Equation~\eqref{pdist}}
    \label{fig:soft-label-pdist-ex}
\end{figure}

In order to teach the student TM to mimic the teacher, the student must be trained with a probability distribution of the teacher's output for each training sample. This probability distribution (see Equation~\eqref{pdist}) determines how likely the input at $p_i$ data belongs to class $i$, with the predicted class being the highest (most likely) value in the distribution.
\begin{equation}\label{pdist}
    \mathbf{p}=\begin{bmatrix}0.0715 & 0.0831 & 0.0872 & 0.1160 & 0.0878 & 0.1040 & 0.0856 & 0.0808 & 0.1940 & 0.0894\end{bmatrix}
\end{equation}
This can example can be visualized in Figure~\ref{fig:soft-label-pdist-ex}. The selected class is shown in red.
Although these distributions are easily obtained in traditional neural networks, they are harder to obtain in Tsetlin Machines. Recall that in a traditional multi-class Tsetlin Machine, a classification decision is made by summing all the positive and negative clauses per class and selecting the class index with the highest output. Let $|C|$ be the number of clauses and $\zeta$ be the number of classes in a Tsetlin Machine. Then Equation~\eqref{eq:wmctm-sum} shows the classification decision calculation for a weighted, multi-class Tsetlin Machine.
\begin{equation}\label{eq:wmctm-sum}
   \hat{y} = \underset{i=1,\dots,\zeta}{\mathrm{argmax}} \left( \sum_{j=1}^{|C|/2} w_j^+C_j^+ - \sum_{j=1}^{|C|/2}w_j^-C_j^-\right)
\end{equation}
In order to create a probability distribution like the one shown in \ref{pdist}, the class sums must be directly obtained before any binary class selection. Class sums $\mathbb{C}\in\mathbb{R}^{\zeta}$ over $\zeta$ classes are computed for each classes shown in Equation~\eqref{eq:csums}:
\begin{equation}
\label{eq:csums}
\mathbb{C} = \begin{bmatrix}
        \sum_{j=1}^{|C|/2} w^+_{j,0}  C_{j,0}^{+} - \sum_{j=1}^{|C|/2} w^-_{j,0} C_{j,0}^{-} \\
        \sum_{j=1}^{|C|/2} w^+_{j,1} C_{j,1}^{+} - \sum_{j=1}^{|C|/2} w^-_{j,1} C_{j,1}^{-} \\
        \vdots \\
        \sum_{j=1}^{|C|/2} w^+_{j,\zeta} C_{j,\zeta}^{+} - \sum_{j=1}^{|C|/2} w^-_{j,\zeta} C_{j,\zeta}^{-} \\
    \end{bmatrix}
\end{equation}
More generally, $\mathbb{C}\in\mathbb{R}^{\zeta}$ can be computed as shown in Equation~\eqref{eq:csums-general}:
\begin{equation}
\label{eq:csums-general}
\mathbb{C}_{k} = \sum_{j=1}^{|C|/2} w^+_{j,k}C_{j,k}^{+} - \sum_{j=1}^{|C|/2}  w^-_{j,k}C_{j,k}^{-}, \quad \forall k \in \{1, \dots, \zeta\}. 
\end{equation}
Note that for the following calculations to be accurate, the class sums must be unclamped. Typical Tsetlin Machine implementations clamp the class sums between $(-T,T)$, which will reduce the information contained in the generated distributions.

\begin{algorithm}
\caption{Generate Soft Labels from Tsetlin Machine Class Sums.}
 \label{algo:soft-labels}
\begin{algorithmic}[1]
\Require $\mathbb{C} \in \mathbb{R}^{N \times \zeta}$: Unclamped class sums for $N$ examples across $\zeta$ classes
\Function{GetSoftLabels}{$\mathbb{C}$}
    \LineComment{Phase 1: Transform class sums to non-negative values}
    \State $c_{\min}^i = \min_{j \in \{1,\ldots,\zeta\}} \mathbb{C}_{ij}, \forall i \in \{1,\ldots,N\}$ \Comment{Find min class sum for each example $i$}
    \State $\hat{\mathbb{C}}_{ij} = \mathbb{C}_{ij} - c_{\min}^i, \forall i,j$ \Comment{Shift values so minimum becomes zero for each example}
    \\
    \LineComment{Phase 2: Scale values to $[0,1]$ interval to enable meaningful comparisons}
    \State $m_i = \max_{j \in \{1,\ldots,\zeta\}} \hat{\mathbb{C}}_{ij}, \forall i \in \{1,\ldots,N\}$ \Comment{Find max shifted value per example}
    \State $\bar{\mathbb{C}}_{ij} = \displaystyle\frac{\hat{\mathbb{C}}_{ij}}{m_i}, \forall i,j \implies \bar{\mathbb{C}}_{ij} \in [0,1]$ \Comment{Normalize to unit interval}
    \\
    \LineComment{Phase 3: Apply softmax to obtain probability distribution over classes}
    \State $P_{ij} = \exp{(\bar{\mathbb{C}}_{ij})}, \forall i,j$ \Comment{Apply exponential function to emphasize larger values}
    \State $S_{ij} = \displaystyle\frac{P_{ij}}{\sum_{k=1}^{\zeta} P_{ik}}, \forall i,j \implies S_{i} \in \Delta^{\zeta-1}$ \Comment{Normalize probabilities} 
    
    \State \Return $S \in \mathbb{R}^{N \times \zeta}$ \Comment{$S_{ij}$ represents $\mathbb{P}(y_i = j | \mathbf{x}_i)$}
\EndFunction
\end{algorithmic}
\end{algorithm}

Once the unclamped class sums have been computed, the normalized probability distributions for each sample are created in Algorithm \ref{algo:soft-labels}. First, the class sums $\mathbb{C}$ are shifted by the absolute value of the minimum sum per sample. This ensures all values are positive. Next, the shifted values $\hat{\mathbb{C}}$ are normalized between $[0,1]$ for each example. The exponential function is applied to every normalized value such that $P_{ij} = \exp{(\bar{\mathbb{C}}_{ij})}, \forall i,j$. Lastly, each sample's normalized and exponentiated class sums undergo a second normalization step $S_{ij} = \displaystyle\frac{P_{ij}}{\sum_{k=1}^{\zeta} P_{ik}}, \forall i,j$, ensuring that the values of every $S_i$ are non-negative and sum to 1. The final soft labels are in space $\mathbb{R}^{N \times \zeta}$, or number of data samples by number of classes, respectively. This is repeated for every input data sample. These soft labels are used in training to help the student mimic the teacher, as explained in the following section.

\subsection{Improved Training}

\begin{figure}
    \centering
    \includegraphics[width=0.85\linewidth]{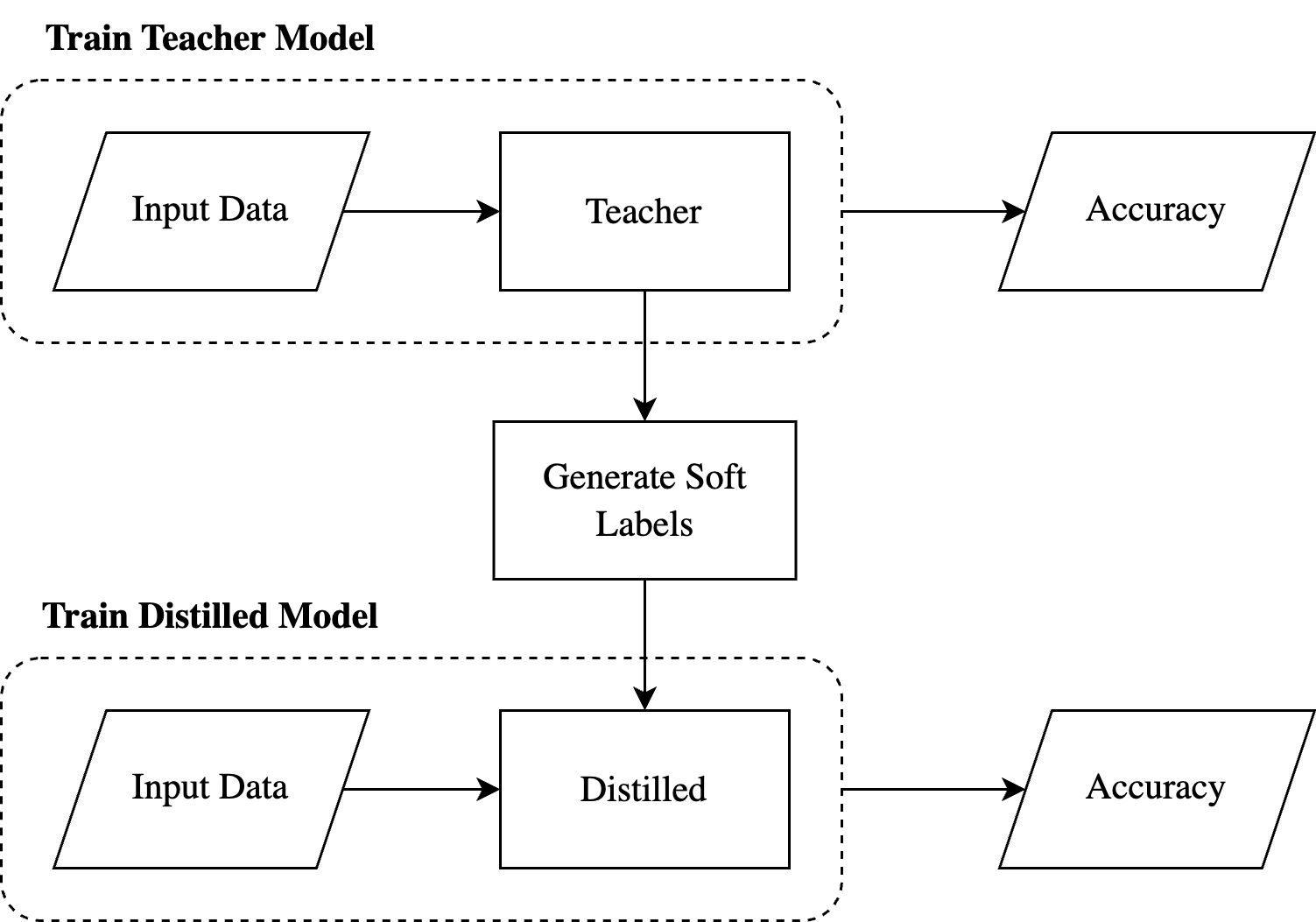}
    \caption{Visualization of the training process using soft labels.}
    \label{fig:soft-labels-flow}
\end{figure}

In order to use the soft labels generated by the teacher in Algorithm~\ref{algo:soft-labels}, we must modify the standard Tsetlin Machine update step. This consists of two phases, reinforcing the positive class and reinforcing the negative classes. Recall that a multi-class Tsetlin Machine is effectively a team of $\zeta$ TMs, one for each class. A typical multi-class TM updates at each fit step by sending Type I feedback to the true class $c$, and sending Type II feedback to a random negative class $\neq c$. We propose a novel fit method for use in Tsetlin Machines in Algorithm \ref{algo:fit-soft}. Like a traditional TM training step, Algorithm \ref{algo:fit-soft} inputs training data $X\in\mathbb{R}^{N \times n}$, true labels $y\in\mathbb{R}^{N}$, number of training examples $N$, and number of epochs $E$. We introduce three new parameters: soft labels $S\in\mathbb{R}^{N \times \zeta}$, balance $\alpha \in [0,1]$, and distribution temperature $\tau \in (0,\infty)$. Balance controls how influential the soft labels are in training, and temperature adjusts the entropy of the soft labels.
Figure~\ref{fig:soft-labels-flow} illustrates the data pathway for training the distilled model using soft labels from the teacher.

\begin{figure}
    \centering
    \includegraphics[width=\linewidth]{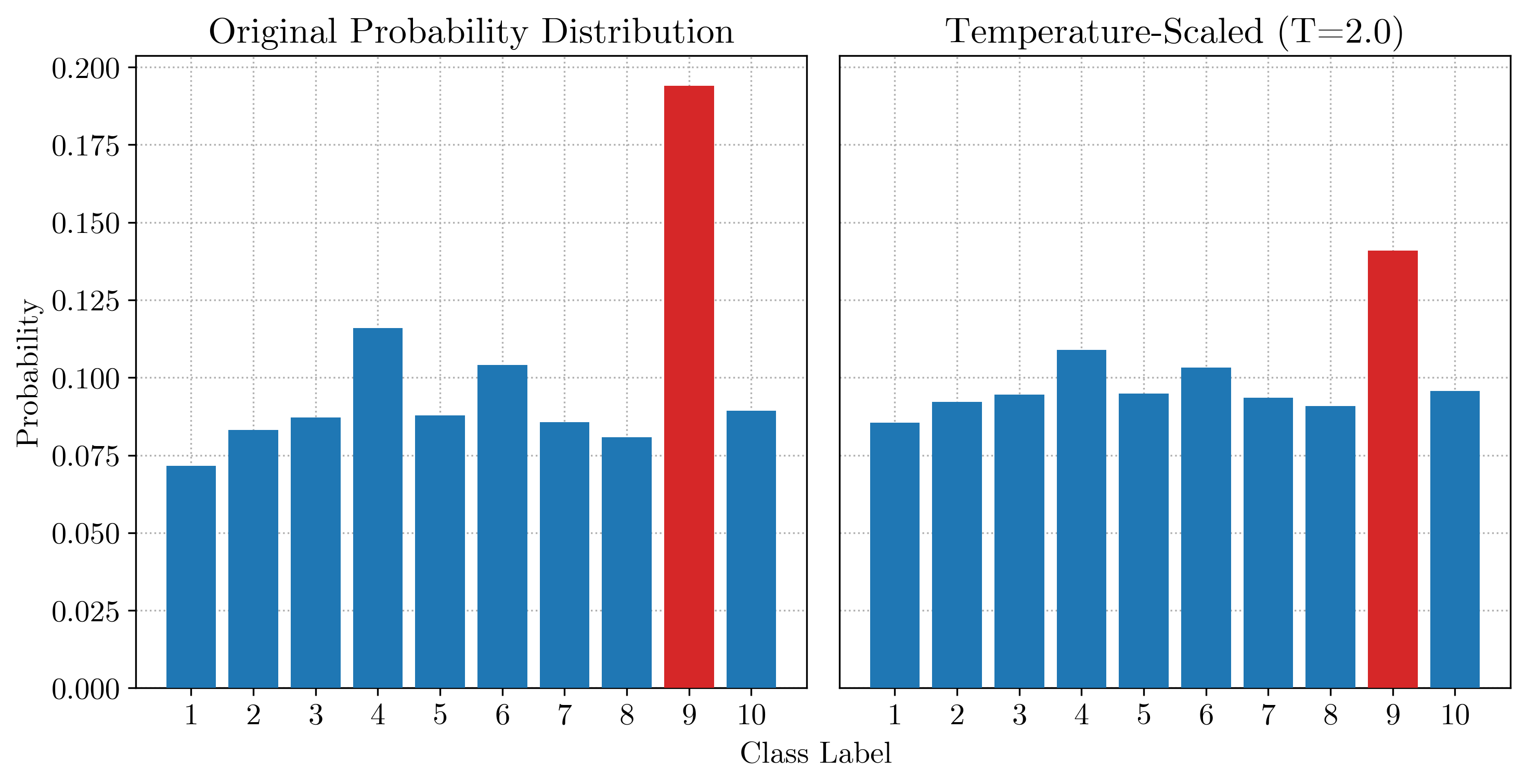}
    \caption{Visualization of the temperature-scaled distribution with selected class highlighted.}
    \label{fig:pdist-ex-scaled}
\end{figure}

First, for each epoch $e$ and example $i$, the soft label distributions are adjusted by temperature $\tau^2$ and re-normalized. This changes the entropy, or uncertainty, of the soft labels. Vanilla softmax can be thought of as adjusted by temperature $\tau=1$. With a $\tau>1$, the entropy is increased, moving the distribution towards the uniform distribution. Setting $\tau<1$ does the opposite, emphasizing the true class. This effect is illustrated in~Figure~\ref{fig:pdist-ex-scaled}. 

Instead of always applying Type I feedback to the correct class $y_i=c$, we apply it with probability $\alpha$. This helps the student learn better from the ambiguous examples in the data and mimic the teacher's behavior. Next, each class $k$ in $\{1,\dots,\zeta\}$, is iterated over. If $k=c$ and $\alpha>0$, the class is skipped as to not provide incorrect feedback to the true class. The base feedback probability $\phi$ for class $k$ is calculated as the $(1-\alpha)p_{i,k}$, or $(1-\alpha)$ times the soft label for the current example and class. Unlikely classes are skipped if $\phi<0.001$ to save computing power. Next, the feedback type for class $k$ is computed. If $p_{i,k} > 0.5$, the soft labels imply that $k$ could be the correct class, so the feedback type is set to Type I feedback. The feedback probability is computed to be $\phi\times(1 + p_{i,k}\tau)$. Otherwise, the feedback type is set to Type II and the feedback probability is set to $\phi\times(1 + (1 - p_{i,k})\tau)$. 
Finally, the selected feedback is applied to the current class with probability $\phi$. This is repeated for each class, example, and epoch.

\begin{algorithm}
\caption{Tsetlin Machine fit method using soft label distribution.}
\label{algo:fit-soft}
\begin{algorithmic}[1]
\Require $X\in\mathbb{R}^{N \times n}$: Training examples, $N$ samples by $n$ features
\Require $y\in\mathbb{R}^{N}$: True class labels vector
\Require $S\in\mathbb{R}^{N \times \zeta}$: Teacher model's probability distributions (soft labels)
\Require $N$: Number of training examples
\Require $E$: Number of epochs
\Require $\alpha \in [0,1]$: Balance between hard/soft labels (0=pure soft, 1=pure hard)
\Require $\tau \in (0,\infty)$: Temperature parameter for sharpening/softening distributions

\Function{FitEnhanced}{$X, y, S, N, E, \alpha, \tau$}
    \For{each epoch $e \in \{1,\ldots,E\}$} \Comment{Iterate through training epochs}
        \For{each example $i \in \{1,\ldots,N\}$} \Comment{Process each training example}
            \State $c \gets y_i$ \Comment{Get true class label for current example}
            
            \State $p_i \gets S_i^{1/\tau^2}$ \Comment{Apply temperature scaling to soften/sharpen distribution}
            \State $p_i \gets \displaystyle\frac{p_i}{\sum p_i}$ \Comment{Renormalize to valid probability distribution}
            \\
            \LineComment{Phase 1: Hard Label Training}
            \If{$u \sim \mathcal{U}(0,1) \leq \alpha$} \Comment{Random variable between 0 and 1}
                \State $\text{TM}_c.\text{update}(X_i, \texttt{Type I})$ \Comment{Train true class with Type I feedback}
            \EndIf
            \\
            \LineComment{Phase 2: Soft Label Training}
            \For{each class $k \in \{1,\ldots,\zeta\}$} \Comment{Process each class}
                \LineComment{Skip if already trained with hard labels}
                \If{$k = c$ \textbf{and} $\alpha > 0$}
                    \State \textbf{continue}
                \EndIf
                
                \LineComment{Calculate base probability for feedback}
                \State $\phi \gets (1 - \alpha)p_{i,k}$ \Comment{Weight soft label by $(1-\alpha)$}
                
                \If{$\phi < 0.001$} \Comment{Skip classes with very low probability}
                    \State \textbf{continue}
                \EndIf

                \LineComment{Determine feedback type and adjust probability}
                \If{$p_{i,k} > 0.5$} \Comment{Only in rare cases is this true}
                    \State $f \gets \texttt{Type I}$ \Comment{Use positive feedback}
                    \State $\phi \gets \phi(1 + p_{i,k}\tau)$ \Comment{Boost probability by confidence}
                \Else \Comment{Teacher thinks this class is unlikely}
                    \State $f \gets \texttt{Type II}$ \Comment{Use negative feedback}
                    \State $\phi \gets \phi(1 + (1 - p_{i,k})\tau)$ \Comment{Boost by inverse confidence}
                \EndIf
                
                \LineComment{Apply feedback probabilistically based on adjusted confidence}
                \If{$u \sim \mathcal{U}(0,1) \leq \phi$} \Comment{Random variable between 0 and 1}
                    \State $\text{TM}_j.\text{update}(X_i, f)$ \Comment{Update TM for class $k$}
                \EndIf
            \EndFor
        \EndFor
    \EndFor
\EndFunction
\end{algorithmic}
\end{algorithm}

 Algorithm~\ref{algo:fit-soft} hijacks the distilled model to learn from the added context of the teacher's output, in addition to the ground truth from the input dataset.

\chapter{Experiment and Results}

In this section, we explore how using a feature- and response-based teacher-student knowledge distillation model can impact accuracy and performance over several different datasets. We test both approaches described in the previous section to determine which approach works best. In both approaches, we compare training and testing accuracy and times to make conclusions on the effectiveness of our process. These experiments are conducted using a parallel Tsetlin Machine implementation~\footnote{https://github.com/ckinateder/pyTsetlinMachineParallel}. 

\section{Datasets}

We use a total of four different standard benchmark datasets for our experiments: MNIST, KMNIST, EMNIST, and IMDB. Numerical descriptions are shown in Table~\ref{tab:dataset-size}.

\subsection{Dataset Descriptions}

\subsubsection{MNIST}
MNIST (Modified National Institute of Standards and Technology)\cite{lecun1998gradient} is a common machine learning dataset with a total of 70,000 samples depicting handwritten digits. The dataset is split into 60,000 training images and 10,000 testing images. Each image is a grayscale digit from $\{0,1,...,9\}$ of size 28x28 pixels. 

\subsubsection{Kuzushiji-MNIST (KMNIST)}
Kuzushiji-MNIST~\cite{kmnist} is another image classification dataset that contains 70,000 grayscale images of historical Japanese characters, with 60,000 training images and 10,000 test images. Each image is 28x28 pixels, and there are 10 classes. This dataset is another alternative to MNIST.

\subsubsection{Extended-MNIST (EMINST)}
EMNIST~\cite{DBLP:EMNIST} is a similar dataset to MNIST, containing 145,600 grayscale 28x28 images. It consists of 124,800 training images and 20,800 testing images spread across 26 classes, one for each letter of the alphabet.

\subsubsection{IMDB}
The IMDB dataset~\cite{maas-EtAl:2011:ACL-HLT2011} is a text classification dataset for sentiment analysis over movie reviews. It contains 25,000 highly polarized movie reviews for training and 25,000 more for testing. There are only two classes: positive and negative. 

\subsection{Dataset Preprocessing}
\begin{figure}
    \centering
    \includegraphics[width=0.84\linewidth]{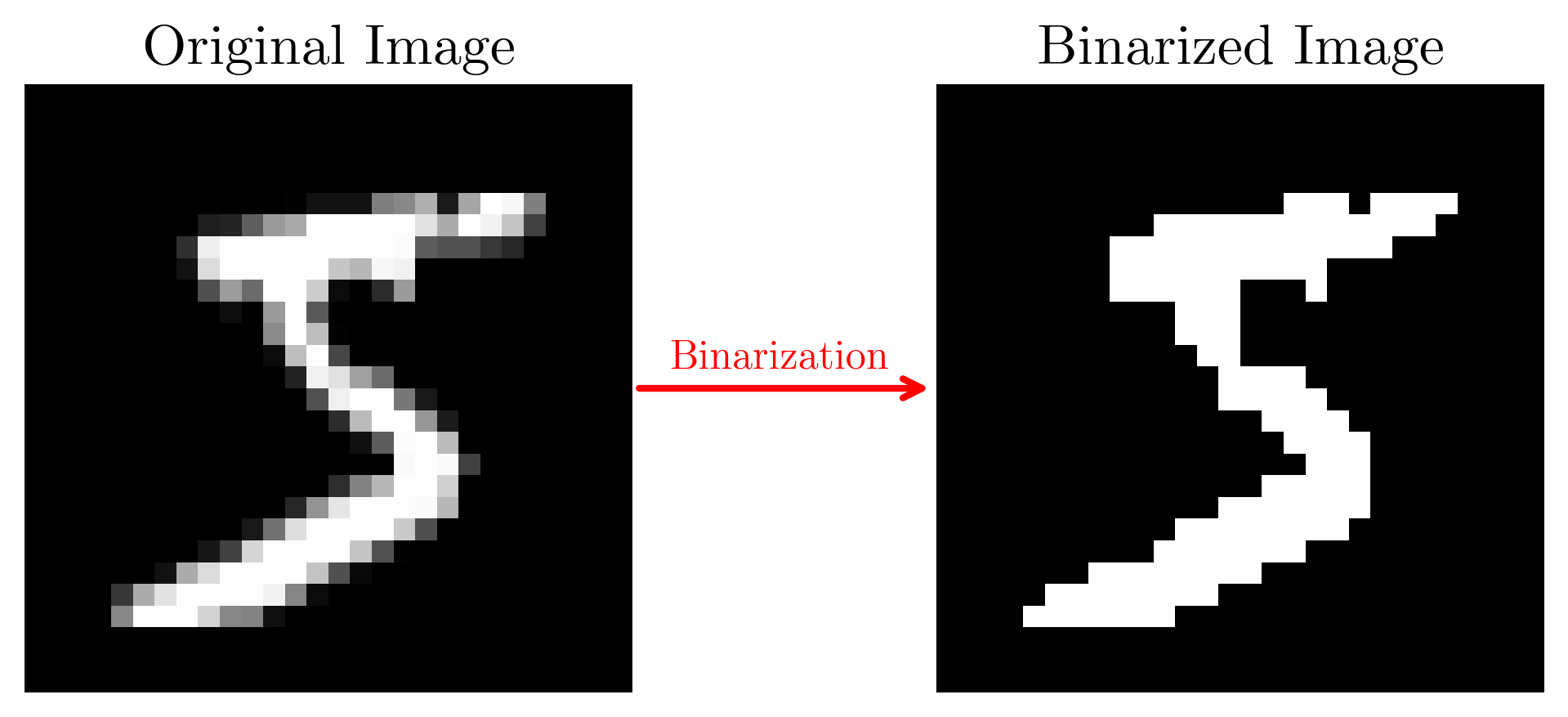}
    \caption{Binarization on an MNIST sample with threshold$=75/255$.}
    \label{fig:binarization}
\end{figure}

For the image experiments, all grayscale image data is binarized at threshold $75/255$ ($0.3$ normalized).
Every pixel with intensity 75 or lower is set to 0, and every pixel above that threshold is set to 1. IMDB textual data is preprocessed into n-grams and decomposed into bit representations before training for a total of 5000 features per sample. This process is shown in Figure~\ref{fig:binarization}.

\begin{table}
\centering
\begin{tabular}{llllll}
\toprule
Dataset & $|X_{train}|$ & $|X_{test}|$ & $|L|$ & $\zeta$ & Type\\
\midrule
EMNIST & 124800 & 20800 & 784 & 26 & Image \\
MNIST & 60000 & 10000 & 784 & 10 & Image \\
KMNIST & 60000 & 10000 & 784 & 10 & Image \\
IMDB & 25000 & 25000 & 5000 & 2 & Text \\
\bottomrule
\end{tabular}
\caption{Dataset Information}
\label{tab:dataset-size}
\end{table}

\section{Experiment Design}

Our experiments are similar to a neural network knowledge distillation scenario. We have a student model $TM_S$, a teacher model $TM_T$, and a distilled model $TM_D$. All three models have the same parameters except for the number of clauses. The student model is assigned clauses than the teacher (in our experiments, between 2 and 10 times fewer), and the distilled model has the same number of clauses as the student model. The student model serves as a baseline for what the distilled model would be without our proposed algorithms. Because the student model and distilled model are the same size, they should be roughly the same speed.

Let $E_S$ and $E_T$ designate the student and teacher epochs, respectively. Let $E_{T+S}$ represent the sum $E_{T+S}=E_T+E_S$.

First, the student model $TM_S$ is trained on the dataset for $E_{T+S}$ epochs. This gives us a baseline of what accuracy can be expected in a Tsetlin Machine with $C_S$ clauses and $E_{T+S}$ epochs. Then, we train the teacher model $TM_T$ with $C_T$ clauses for $E_{T+S}$. 

\paragraph{Clause-Based Knowledge Distillation (CKD).} Once baselines are established, we reinitialize the teacher model $TM_T$ from the checkpoint at $E_T$ epochs to avoid giving the distilled model an unfair advantage. Using the teacher model. We then generate the clause outputs over both the training and testing dataset portions. This is done with the mechanism described in Section~\ref{sec:clause-output}. We use this output to train the distilled model $TM_D$ for $E_S$ epochs. We also apply Probabilistic Clause Downsampling Algorithm~\ref{algo:pcd}) to the clause output and train the downsampled model $TM_{PCD}$ over the downsampled data for $E_S$ epochs. The downsample rate $\delta$ for PCD is chosen to find the best compromise between accuracy and speed. Results are not aggregated with this approach, as CKD is included to show our incremental progress towards DKD and is not the chief focus of this paper.
\paragraph{Distribution-Enhanced Knowledge Distillation (DKD).} After the baseline training is completed, we reinitialize the teacher model $TM_T$ in the same way described for CKD. Next, we initialize the distilled model $TM_D$ with the clauses of the trained teacher, using Algorithm \ref{algo:init-clauses}. We then generate the soft labels from the teacher over the training dataset using Algorithm \ref{algo:soft-labels}. Finally, we train the distilled model for $E_S$ epochs on the training dataset and the soft labels from the teacher using Algorithm \ref{algo:fit-soft}. Due to the stochastic nature of Tsetlin Machines, we run the experiment $K$ times and average the results.

For data collection, we measure all average accuracies over $E_{T+S}$ epochs. We measure the training and testing times of the student and teacher baselines over $E_{T+S}$ epochs, and the distilled training and testing times over $E_S$ epochs. This reflects the design that the distilled model starts its training after $E_T$ epochs with a smaller architecture than the teacher. Measuring distilled average epoch time over $E_{T+S}$ epochs would incorrectly skew the measurements. The steps to generate soft labels, generate clause outputs, and apply PCD are not timed --- this is a one time cost approximately equal to the time of one training epoch.

\section{Clause-Based KD Results}

\begin{table}
\centering
\begin{tabular}{llllllllll}
\toprule
Dataset & $|C_T|$ & $|C_S|$ & $T_T$ & $T_S$ & $s_T$ & $s_S$ & $\delta^*$ & $E_T$ & $E_S$ \\
\midrule
MNIST & 800 & 100 & 10 & 10 & 7.0 & 7.0 & 0.15 & 120 & 240 \\
KMNIST & 400 & 100 & 100 & 100 & 5 & 5 & 0.22 & 120 & 240 \\
EMNIST & 400 & 100 & 100 & 100 & 4.0 & 4.0 & 0.25 & 120 & 240 \\
IMDB & 10000 & 2000 & 6000 & 6000 & 5.0 & 5.0 & 0.15 & 30 & 90 \\
\bottomrule
\end{tabular}
\caption{Experiment Hyperparameters (CKD)}
\label{tab:hyperparams-ckd}
\end{table}

Hyperparameters for our clause-based experiments are detailed in Table~\ref{tab:hyperparams-ckd}.

\subsection{Training Accuracy and Performance Analysis} 

\begin{table}
\centering
\begin{tabular}{lCCCCCCCC}
\toprule
Dataset & $Acc'_T$ & $\mathcal{T}'_T$ & $Acc'_S$ & $\mathcal{T}'_S$ & $Acc'_D$ & $\mathcal{T}'_D$ & $Acc'_{PCD}$ & $\mathcal{T}'_{PCD}$ \\
\midrule
MNIST & 94.80 $\pm$ 1.15 & 0.62 & 91.19 $\pm$ 1.10 & 0.25 & 94.29 $\pm$ 1.20 & 1.46 & 92.40 $\pm$ 1.19 & 0.39 \\
KMNIST & 96.63 $\pm$ 1.60 & 0.44 & 90.61 $\pm$ 0.94 & 0.30 & 96.13 $\pm$ 1.72 & 0.88 & 92.32 $\pm$ 1.06 & 0.38 \\
EMNIST & 85.60 $\pm$ 3.01 & 1.04 & 79.36 $\pm$ 1.50 & 0.62 & 86.10 $\pm$ 3.61 & 5.39 & 83.08 $\pm$ 2.47 & 1.79 \\
IMDB & 99.69 $\pm$ 1.20 & 36.78 & 97.68 $\pm$ 1.92 & 7.72 & 99.39 $\pm$ 1.31 & 30.61 & 99.43 $\pm$ 1.42 & 24.84 \\
\bottomrule
\end{tabular}
\caption{Training Results (CKD)}
\label{tab:train-table-ckd}
\end{table}

\begin{figure}
    \centering
    \includegraphics[width=\linewidth]{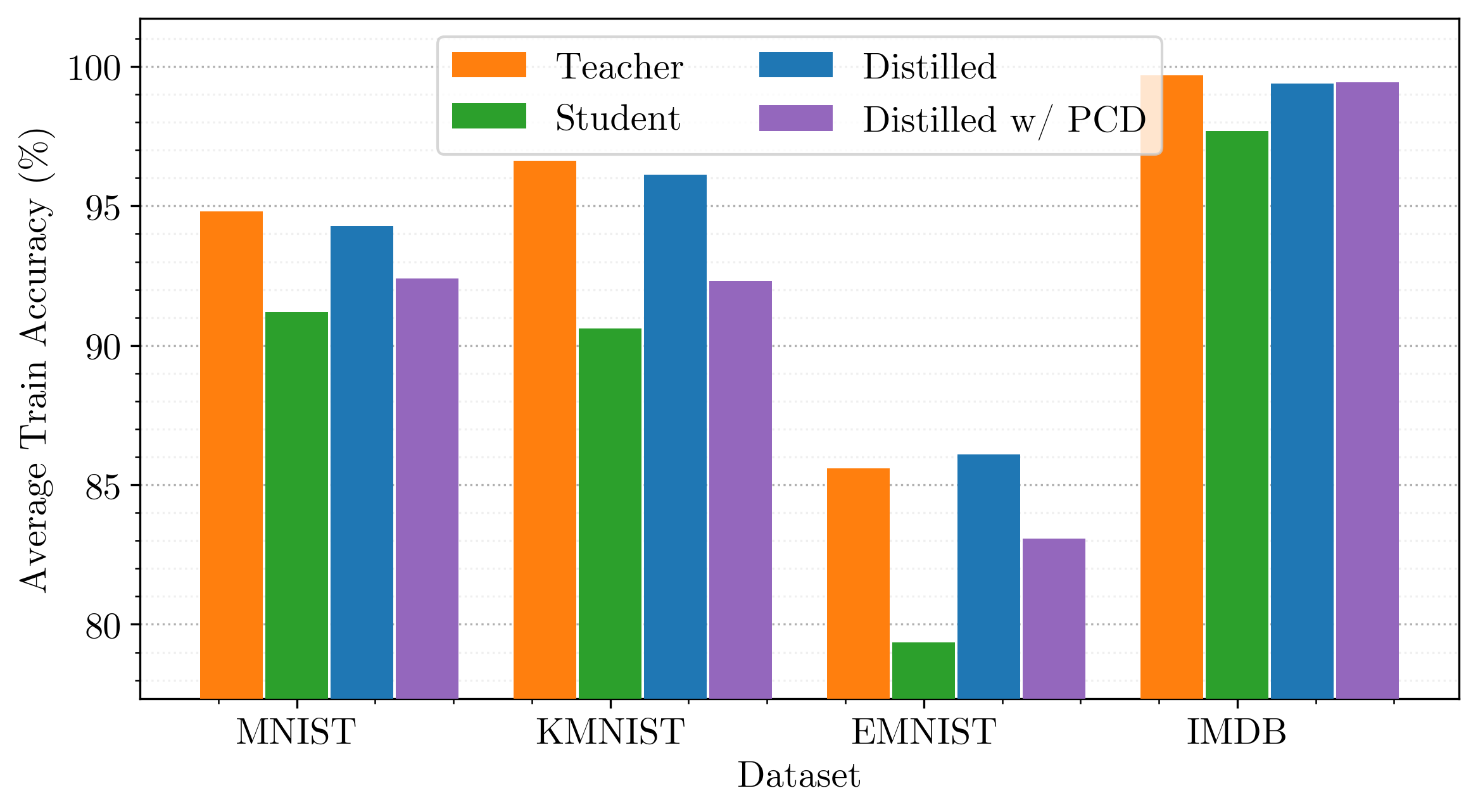}
    \caption{Average training accuracy $Acc'$ on each dataset using CKD.}
    \label{fig:train-acc-combined-ckd}
\end{figure}

We observe for all datasets a very minor benefit in training. Figure~\ref{fig:train-acc-combined-ckd} visualizes the accuracies given in \ref{tab:train-table-ckd}. We see that the clause-based distilled model almost reaches the teacher's accuracy in all cases, and the distilled model using PCD sets the accuracy somewhere in between the baseline teacher and student. 

\begin{figure}
    \centering
    \includegraphics[width=\linewidth]{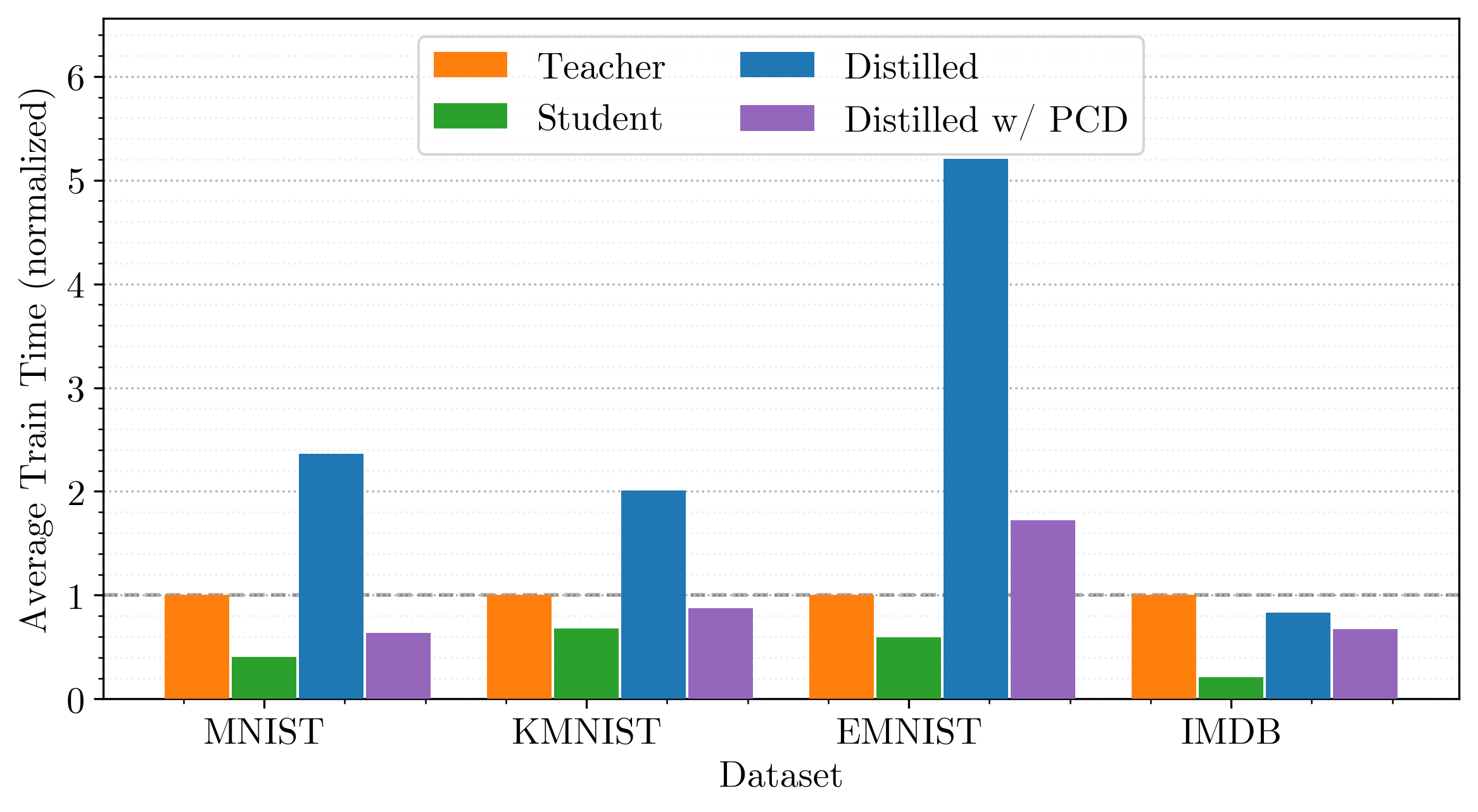}
    \caption{Average epoch training time $\mathcal{T}'$ on each dataset using CKD (relative to teacher).}
    \label{fig:train-time-combined-ckd}
\end{figure}

The main downside to the CKD method is illustrated in Figure~\ref{fig:train-time-combined-ckd}. Because of the increased number of input features in the distilled model $TM_D$, the training time is greatly increased when compared to the student model. Using PCD on the distilled and downsampled model $TM_{PCD}$ does mitigate this, but not always in a worthwhile way. These results combined with Equation~\eqref{eq:literals-d} support the statement that CKD with PCD is best used in datasets with a lower number of classes, like the IMDB dataset.

Examining the training accuracy for the EMNIST dataset shows the distilled model's accuracy at a noticeably higher average than the original teacher model. However, the $TM_{PCD}$ accuracy is between the teacher and the student, but with a much higher training time than the teacher. This suggests that CKD can also be used to boost the teacher's accuracy, but with a greater time cost.
\begin{figure}
    \centering
    \includegraphics[width=\linewidth]{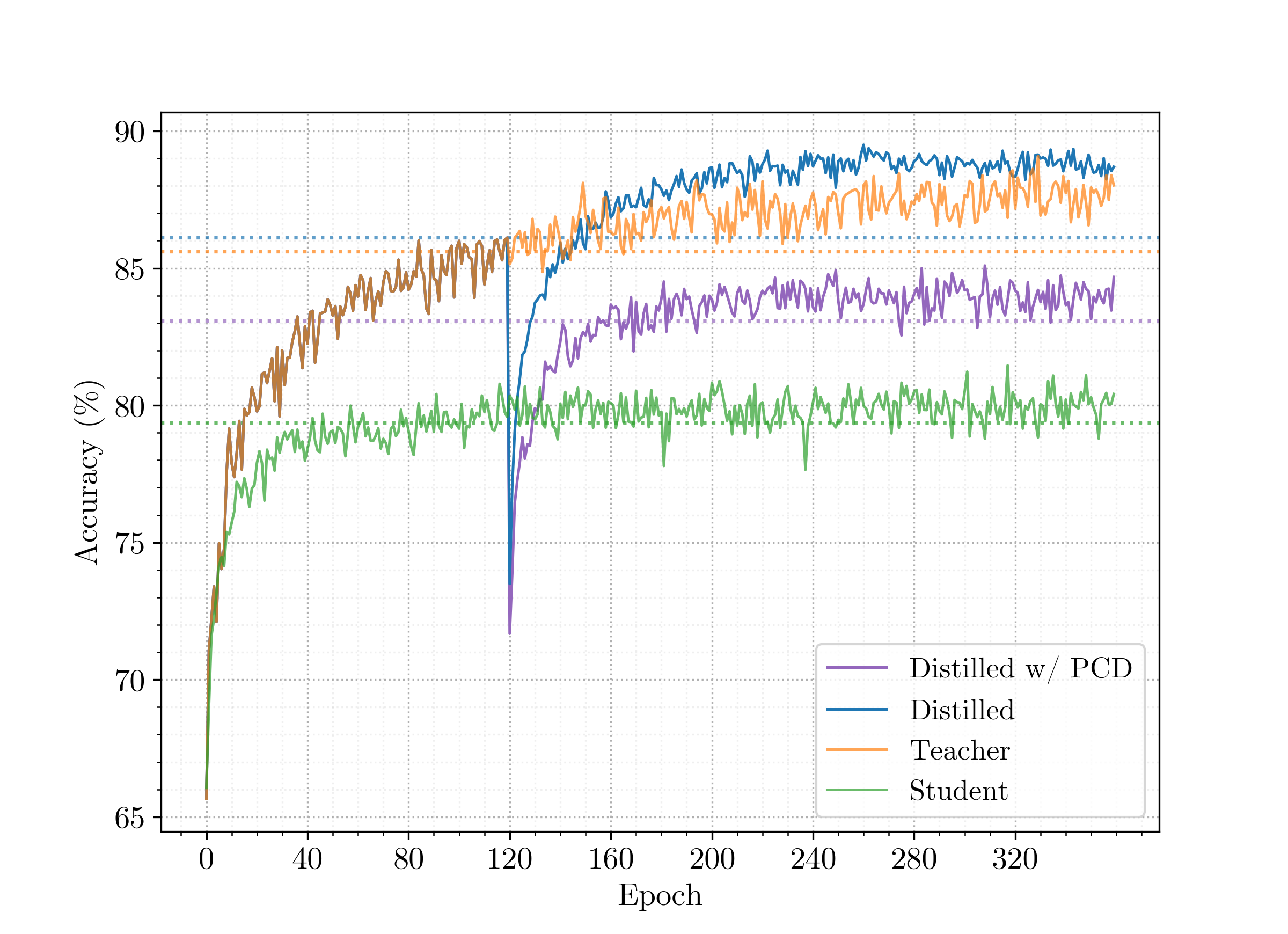}
    \caption{Training accuracy $Acc'$ on the EMNIST dataset using CKD.}
    \label{fig:train-acc-emnist-ckd}
\end{figure}

MNIST and KMNIST both show similar results. Their distilled model training times are higher than the teacher's, but the downsampled training times are reduced. The time it takes to create the clause outputs from the teacher model is negligible as it only needs to happen once for the whole training cycle.

\subsection{Testing Accuracy and Performance Analysis} 

\begin{table}
\centering
\begin{tabular}{lCCCCCCCC}
\toprule
Dataset & $Acc_T$ & $\mathcal{T}_T$ & $Acc_S$ & $\mathcal{T}_S$ & $Acc_D$ & $\mathcal{T}_D$ & $Acc_{PCD}$ & $\mathcal{T}_{PCD}$ \\
\midrule
MNIST & 93.55 $\pm$ 0.95 & 0.29 & 90.90 $\pm$ 1.05 & 0.05 & 93.40 $\pm$ 1.07 & 0.55 & 91.91 $\pm$ 1.11 & 0.10 \\
KMNIST & 84.32 $\pm$ 1.68 & 0.18 & 78.08 $\pm$ 1.26 & 0.06 & 84.04 $\pm$ 1.91 & 0.26 & 79.88 $\pm$ 1.36 & 0.08 \\
EMNIST & 80.82 $\pm$ 2.07 & 0.93 & 77.63 $\pm$ 1.32 & 0.26 & 82.26 $\pm$ 2.99 & 2.35 & 80.44 $\pm$ 2.14 & 0.74 \\
IMDB & 89.00 $\pm$ 0.23 & 23.26 & 87.59 $\pm$ 1.34 & 4.89 & 88.51 $\pm$ 0.52 & 23.92 & 88.76 $\pm$ 0.28 & 19.36 \\
\bottomrule
\end{tabular}
\caption{Testing Results (CKD)}
\label{tab:test-table-ckd}
\end{table}

\begin{figure}
    \centering
    \includegraphics[width=\linewidth]{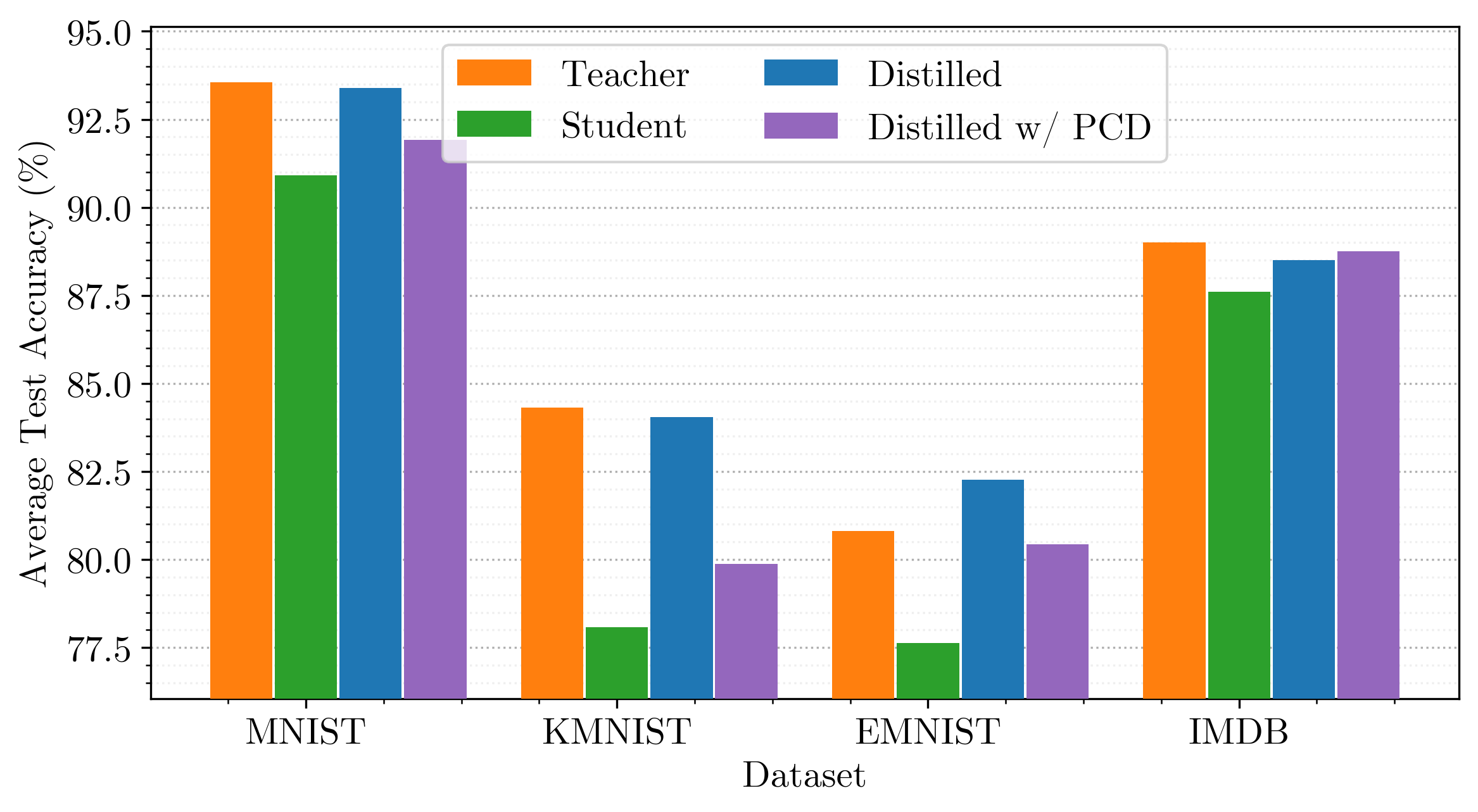}
    \caption{Average testing accuracy $Acc$ on each dataset using CKD.}
    \label{fig:test-acc-combined-ckd}
\end{figure}

\begin{figure}
    \centering
    \includegraphics[width=\linewidth]{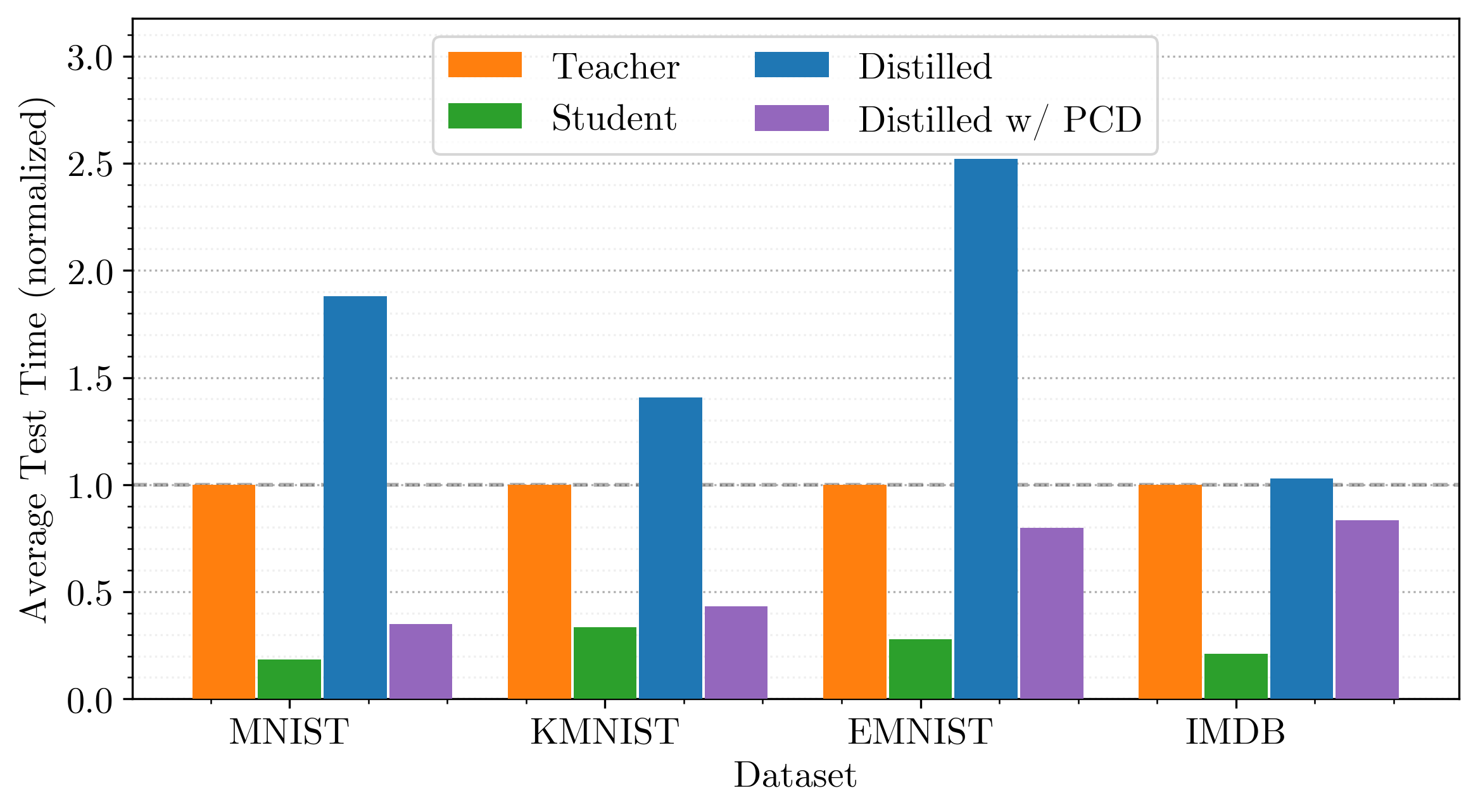}
    \caption{Average epoch testing time $\mathcal{T}$ on each dataset using CKD (relative to teacher).}
    \label{fig:test-time-combined-ckd}
\end{figure}

Testing accuracy and performance results for the CKD experiments are shown in Table~\ref{tab:test-table-ckd}. 
Testing accuracy across all datasets is illustrated in Figure \ref{fig:test-acc-combined-ckd}, and average testing epoch time is illustrated in Figure \ref{fig:test-time-combined-ckd}. 


As stated earlier, the time to generate the clause outputs from the teacher that are fed to the distilled models here is equal to the time it takes to run inference on the teacher model and is not included in Figure~\ref{fig:test-time-combined-ckd}. These outputs must be generated for any unseen data. This is inconsequential in training, as every training epoch occurs on the same data. However, classification models are designed to eventually be used on unseen input data to return a prediction when the ground truth is not known. This happens in a 1:1~ratio~---~new data is typically only seen once. Since the data input to the distilled models must first pass through the teacher model, the total time (including clause generation) for the distilled model equals $\mathcal{T_T}+\mathcal{T}_D$. Therefore, the distilled model can never be faster than the teacher model on unseen data.

\begin{figure}
    \centering
    \includegraphics[width=\linewidth]{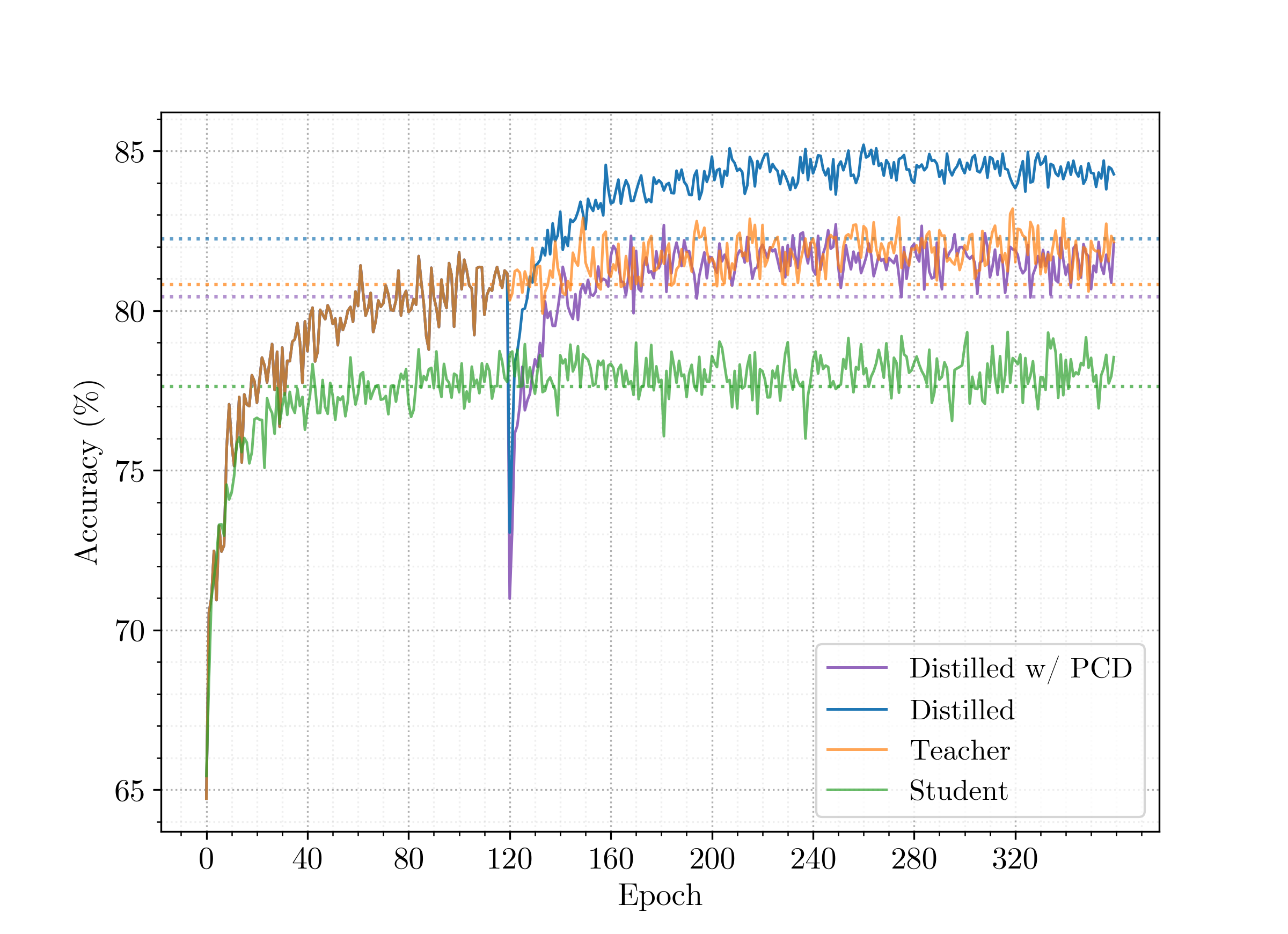}
    \caption{Testing accuracy $Acc$ on the EMNIST dataset using CKD.}
    \label{fig:test-acc-emnist-ckd}
\end{figure}

However, there can still be an accuracy boost. Figure~\ref{fig:test-acc-emnist-ckd} shows the testing accuracy per epoch over the EMNIST dataset. The average accuracy of the distilled model is significantly higher than the teacher model. Additionally, the distilled model's average accuracy is far lower than the final accuracy, suggesting that with more epochs, the average accuracy can still increase. Unfortunately, that accuracy comes at the price of execution time.

\section{Distribution-Enhanced KD Results}
\begin{table}
\centering
\begin{tabular}{lllllllllllll}
\toprule
Dataset & $|C_T|$ & $|C_S|$ & $T_T$ & $T_S$ & $s_T$ & $s_S$ & $\tau$ & $\alpha$ & $z$ & $E_T$ & $E_S$ & $K$ \\
\midrule
EMNIST & 1000 & 100 & 100 & 100 & 4.0 & 4.0 & 4.0 & 0.5 & 0.2 & 120 & 240 & 10 \\
MNIST & 1000 & 100 & 10 & 10 & 4.0 & 4.0 & 3.0 & 0.5 & 0.3 & 120 & 240 & 10 \\
KMNIST & 2000 & 200 & 100 & 100 & 8.2 & 8.2 & 4.0 & 0.5 & 0.3 & 120 & 240 & 10 \\
IMDB & 8000 & 4000 & 6000 & 6000 & 7.0 & 7.0 & 3.0 & 0.5 & 0.2 & 30 & 60 & 8 \\
\bottomrule
\end{tabular}
\caption{Experiment Hyperparameters (DKD)}
\label{tab:hyperparams-dkd}
\end{table}

The hyperparameters for our distribution-enhanced experiments are shown in Table \ref{tab:hyperparams-dkd}.
We observe over all datasets that using our proposed novel knowledge distillation approach has a significant impact on accuracy and performance in the distilled model. Additionally, we observe that the effect of the teacher on the distilled model is visible in the included clause literals.

Tables \ref{tab:train-table-dkd} and \ref{tab:test-table-dkd} report on accuracy and performance over the training and testing datasets, respectively.

\subsection{Training Accuracy and Performance Analysis} \label{sec:trapa}

\begin{table}
\centering
\begin{tabular}{lCCCCCC}
\toprule
Dataset & $Acc'_T$ & $\mathcal{T}'_T$ & $Acc'_S$ & $\mathcal{T}'_S$ & $Acc'_D$ & $\mathcal{T}'_D$ \\
\midrule
MNIST & 97.65 \newline $\pm$ 0.07 & 0.70 \newline $\pm$ 0.01 & 90.63 \newline $\pm$ 0.13 & 0.26 \newline $\pm$ 0.00 & 95.25 \newline $\pm$ 0.07 & 0.24 \newline $\pm$ 0.00 \\
KMNIST & 98.10 \newline $\pm$ 0.03 & 1.69 \newline $\pm$ 0.05 & 92.46 \newline $\pm$ 0.03 & 0.34 \newline $\pm$ 0.00 & 94.66 \newline $\pm$ 0.04 & 0.36 \newline $\pm$ 0.00 \\
EMNIST & 86.88 \newline $\pm$ 0.09 & 3.15 \newline $\pm$ 0.02 & 79.35 \newline $\pm$ 0.06 & 0.62 \newline $\pm$ 0.00 & 83.20 \newline $\pm$ 0.06 & 0.67 \newline $\pm$ 0.00 \\
IMDB & 99.41 \newline $\pm$ 0.07 & 28.95 \newline $\pm$ 0.42 & 98.39 \newline $\pm$ 0.24 & 17.07 \newline $\pm$ 0.35 & 98.96 \newline $\pm$ 0.13 & 9.35 \newline $\pm$ 0.29 \\
\bottomrule
\end{tabular}
\caption{Training Results (DKD)}
\label{tab:train-table-dkd}
\end{table}

\begin{figure}
    \centering
    \includegraphics[width=\linewidth]{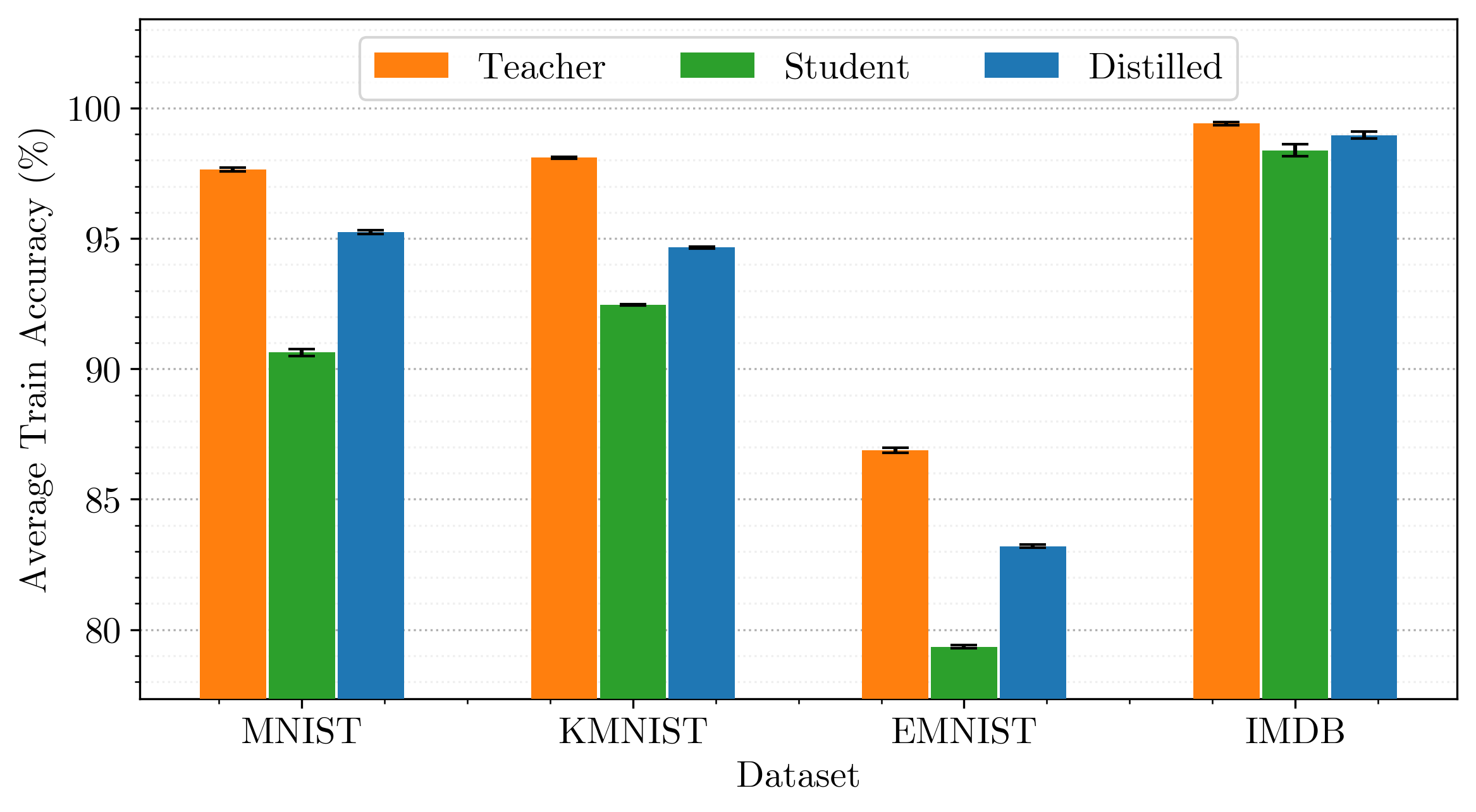}
    \caption{Average training accuracy $Acc'$ on each dataset using DKD.}
    \label{fig:train-acc-combined}
\end{figure}
\begin{figure}
    \centering
    \includegraphics[width=\linewidth]{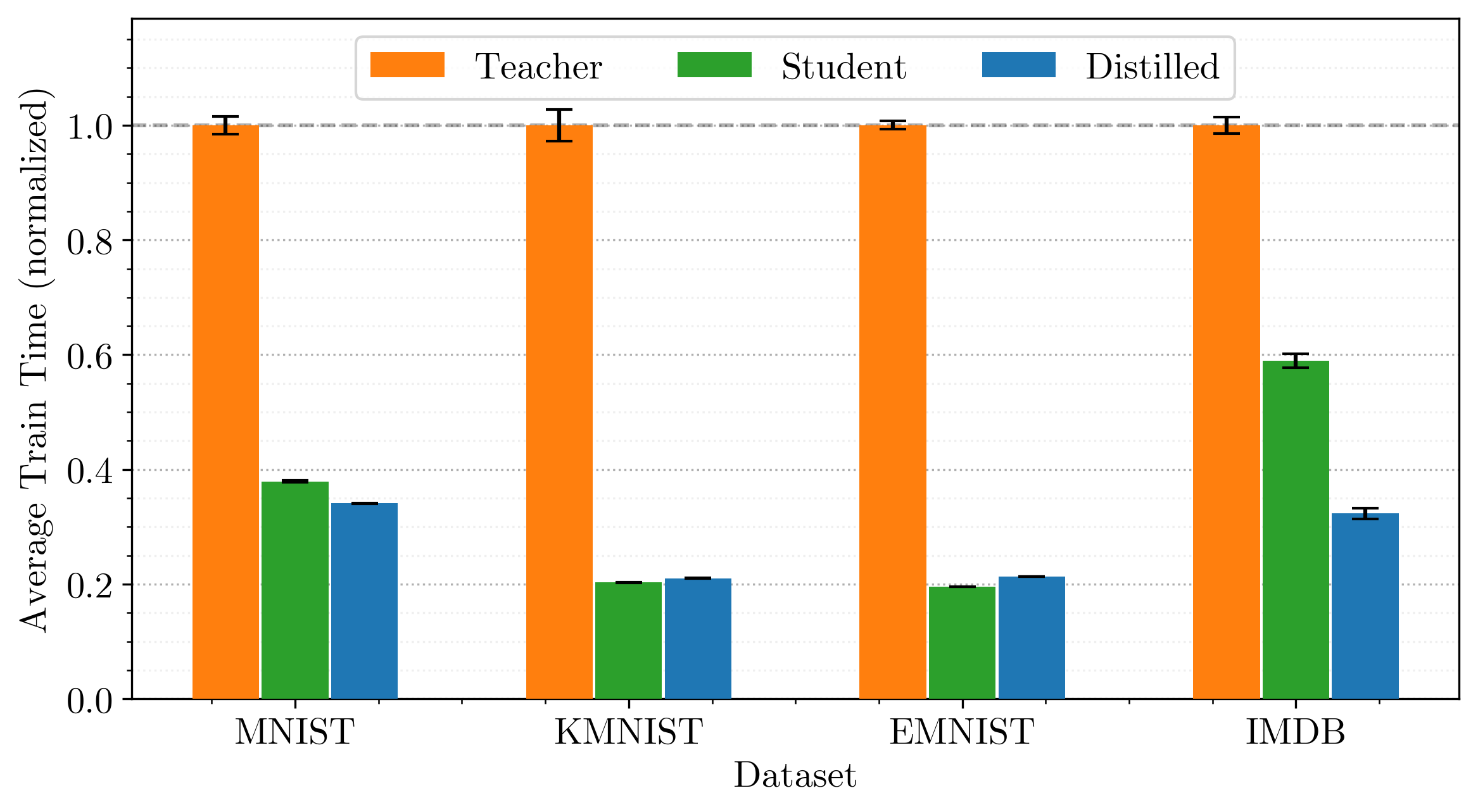}
    \caption{Average epoch training time $\mathcal{T}'$ on each dataset using DKD.}
    \label{fig:train-time-combined}
\end{figure}
\begin{figure}
    \centering
    \includegraphics[width=\linewidth]{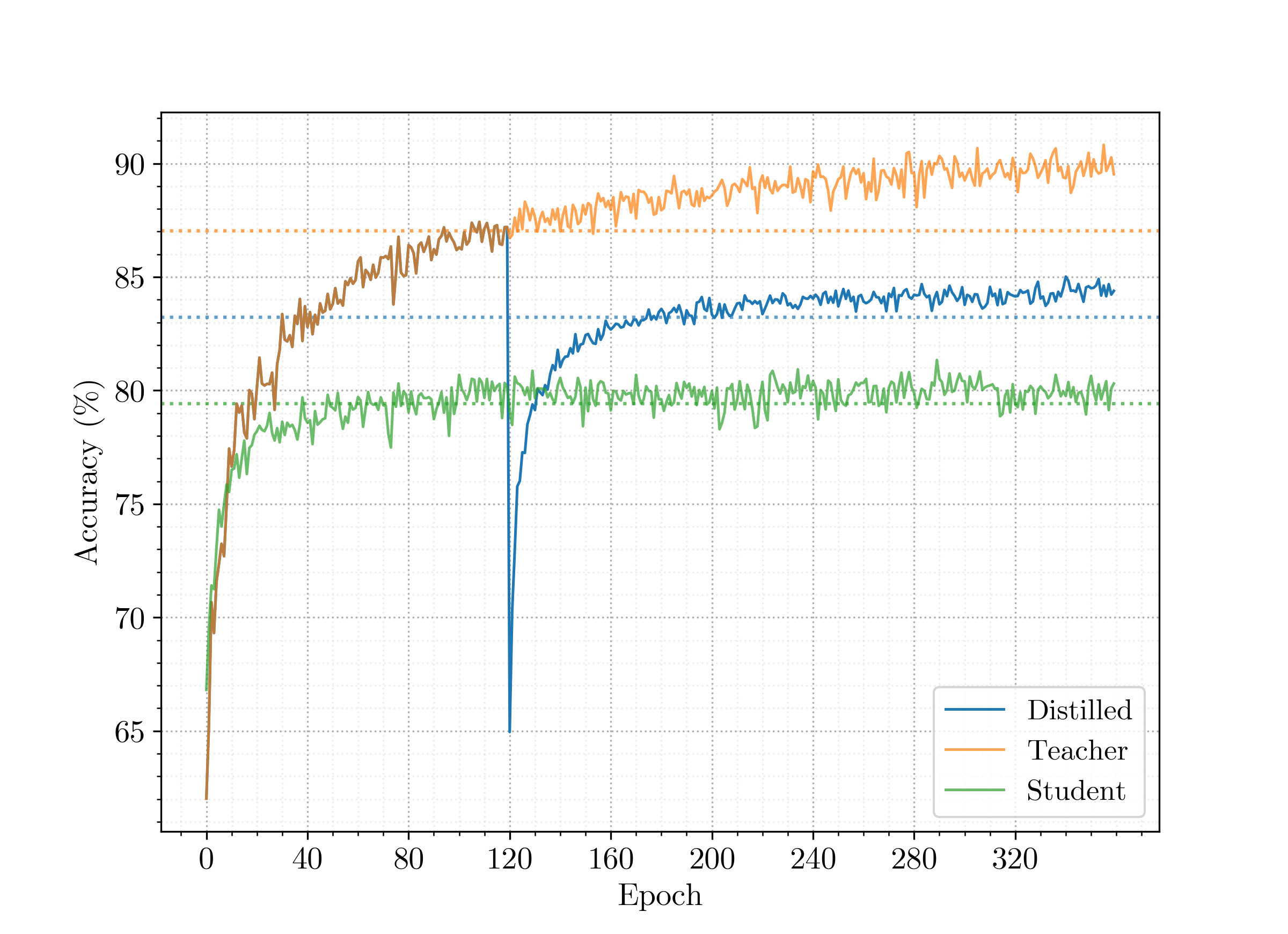}
    \caption{Training accuracy $Acc'$ on the EMNIST dataset using DKD.}
    \label{fig:train-acc-emnist}
\end{figure}
\begin{figure}
    \centering
    \includegraphics[width=\linewidth]{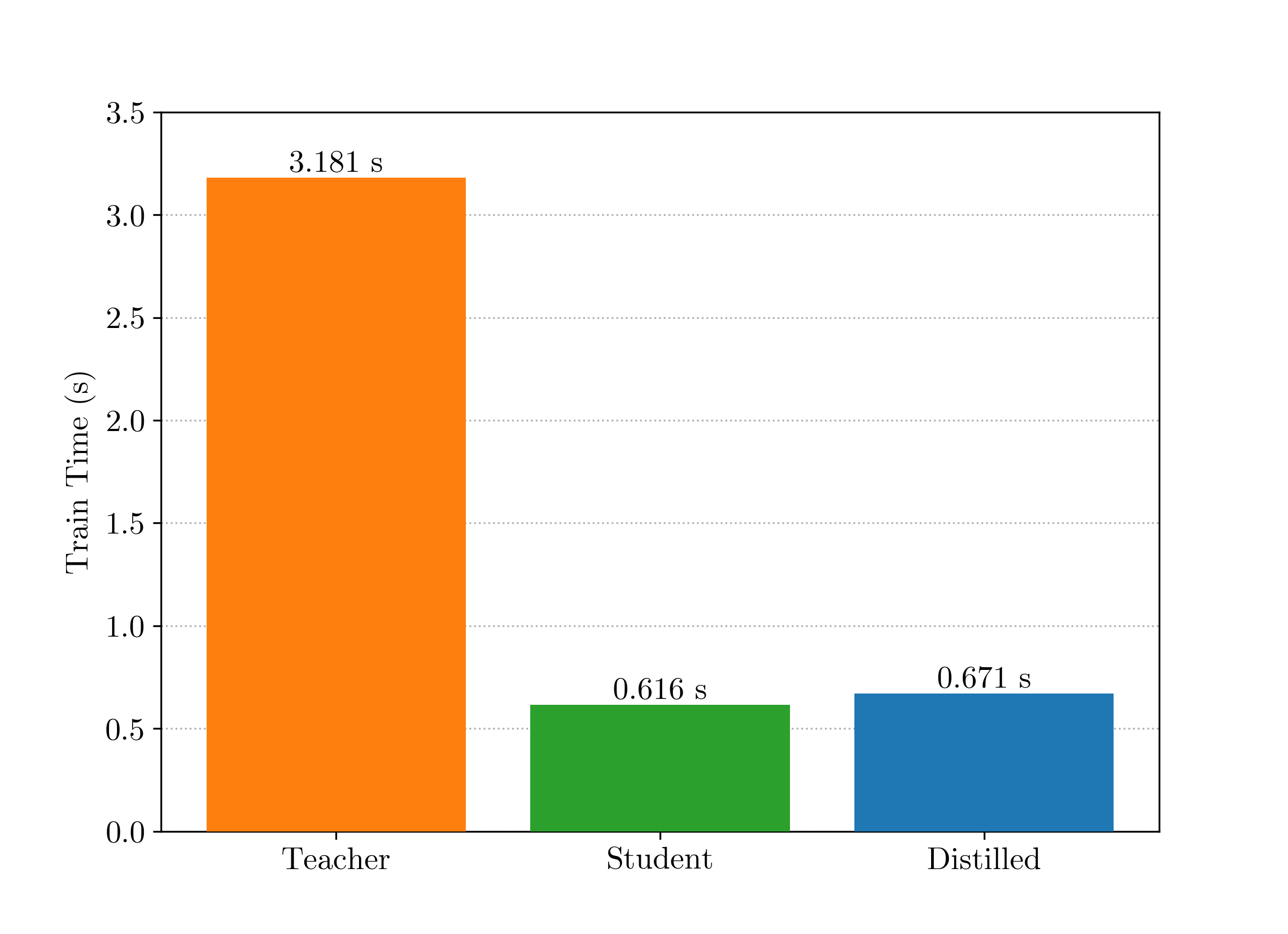}
    \caption{Average training time $\mathcal{T}'$ on the EMNIST dataset using DKD.}
    \label{fig:train-time-emnist}
\end{figure}

Experimental results suggest that using a teacher-student knowledge distillation models can greatly improve the training accuracy of a smaller model with minimal overhead. Initial clause transfer between the teacher and the distilled model, combined with distribution-enhanced fitting, enables the distilled model to considerably increase its accuracy without increasing training time. These results are shown in Table~\ref{tab:train-table-dkd}, Figure~\ref{fig:train-acc-combined}, and Figure~\ref{fig:train-time-combined}.
Note that the average training times are normalized for each experiment in Figure~\ref{fig:train-time-combined}. Normalization was performed to illustrate what the execution time would be for each model if each dataset was the same size.
Over the EMNIST dataset, the training accuracy is improved by almost 4 percentage points when compared to the baseline student, with a similar evaluation time per epoch. This is illustrated in Figure~\ref{fig:train-acc-emnist} and Figure~\ref{fig:train-time-emnist}. 

\begin{figure}
    \centering
    \includegraphics[width=\linewidth]{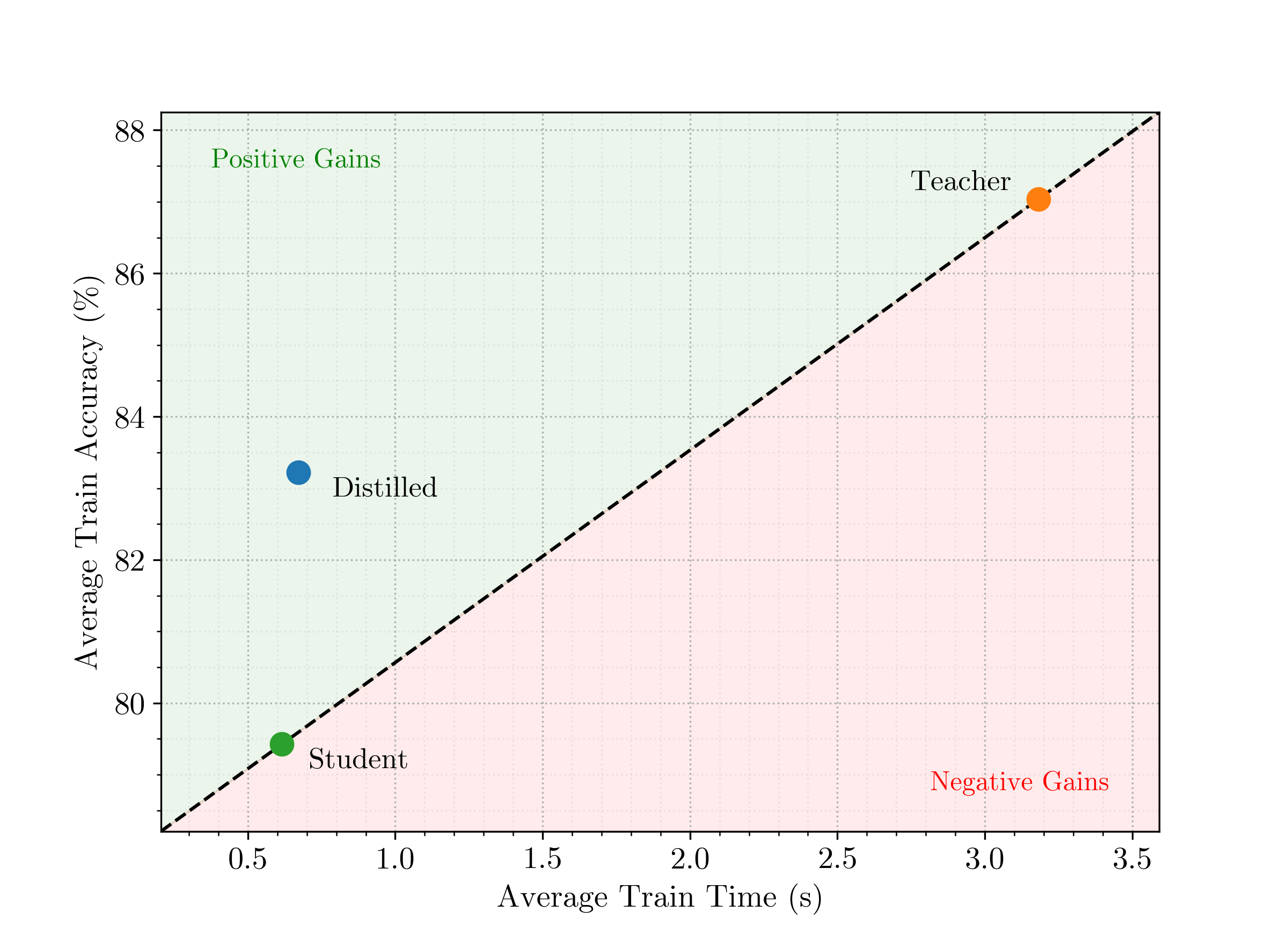}
    \caption{Training efficiency on the EMNIST dataset using DKD.}
    \label{fig:train-eff-emnist}
\end{figure}

The relationship between training time and accuracy can also be visualized in an efficiency plot (Figure~\ref{fig:train-eff-emnist}). Points to the left of the line represent a ``free lunch'', an area in which there is an increase in accuracy without a linear increase in execution time. This graph captures the central idea of knowledge distillation – increasing accuracy without increasing execution time.

Performance on the MNIST dataset is similar to EMNIST. The distilled model $Acc_D$ increases training accuracy over the student by around 5 percentage points, with a significant decrease in variance and a slight decrease in training time. 
The distilled model $TM_D$ trained on the KMNIST dataset sees modest performance gains when compared to the student. Average training accuracy is increased by approximately 2 percentage points, and average epoch time is about the same. 

Data on the IMDB dataset shows an even better result. $TM_D$'s accuracy is in the middle between the teacher and the student, but training time $\mathcal{T}'_D$ is around half of $\mathcal{T}'_S$. This is possible due to the literals that are dropped quickly by the context of the teacher. We also found that IMDB's results stabilized more quickly than any of the MNIST datasets, hence the epoch reduction.

The temporary drop in accuracy at $E=E_T$ shown in each accuracy plot is indicative of the distilled model's need to balance the clauses copied over from the teacher. Recall that $TM_T$ is trained for $E_T$ epochs, and then $|C_S|$ clauses are selected from $TM_T$ and copied to $TM_D$. In the teacher, those selected clauses are used for classification in conjunction with all the $|C_T|-|C_S|$ other clauses. The distilled model does not have those other clauses to factor in, so it must quickly adjust. For example, in Figure~\ref{fig:train-acc-emnist}, it takes the distilled model around 7 epochs to reach the equivalent equilibrium accuracy of the student, a shorter amount of time than the baseline student takes to reach when training from scratch. This suggests that, when a teacher model already exists, clause initialization combined with distribution-enhanced fitting is an optimal choice for training a smaller model in a handful of epochs.

\subsection{Testing Accuracy and Performance Analysis}

The findings over the test datasets further support our findings that accuracy and performance can be improved with our teacher-student framework. Table~\ref{tab:test-table-dkd} shows the performance of each model over the testing datasets. 

\begin{table}
\centering
\begin{tabular}{lCCCCCC}
\toprule
Dataset & $Acc_T$ & $\mathcal{T}_T$ & $Acc_S$ & $\mathcal{T}_S$ & $Acc_D$ & $\mathcal{T}_D$ \\
\midrule
MNIST & 95.20 \newline $\pm$ 0.06 & 0.37 \newline $\pm$ 0.01 & 90.37 \newline $\pm$ 0.14 & 0.06 \newline $\pm$ 0.00 & 93.96 \newline $\pm$ 0.06 & 0.06 \newline $\pm$ 0.00 \\
KMNIST & 85.37 \newline $\pm$ 0.10 & 0.97 \newline $\pm$ 0.01 & 80.11 \newline $\pm$ 0.10 & 0.10 \newline $\pm$ 0.00 & 82.48 \newline $\pm$ 0.07 & 0.10 \newline $\pm$ 0.00 \\
EMNIST & 80.96 \newline $\pm$ 0.10 & 3.27 \newline $\pm$ 0.06 & 77.52 \newline $\pm$ 0.05 & 0.26 \newline $\pm$ 0.00 & 80.31 \newline $\pm$ 0.07 & 0.26 \newline $\pm$ 0.00 \\
IMDB & 88.96 \newline $\pm$ 0.04 & 17.88 \newline $\pm$ 0.71 & 87.85 \newline $\pm$ 0.24 & 10.37 \newline $\pm$ 0.30 & 88.62 \newline $\pm$ 0.03 & 9.36 \newline $\pm$ 0.17 \\
\bottomrule
\end{tabular}
\caption{Testing Results (DKD)}
\label{tab:test-table-dkd}
\end{table}

\begin{figure}
    \centering
    \includegraphics[width=\linewidth]{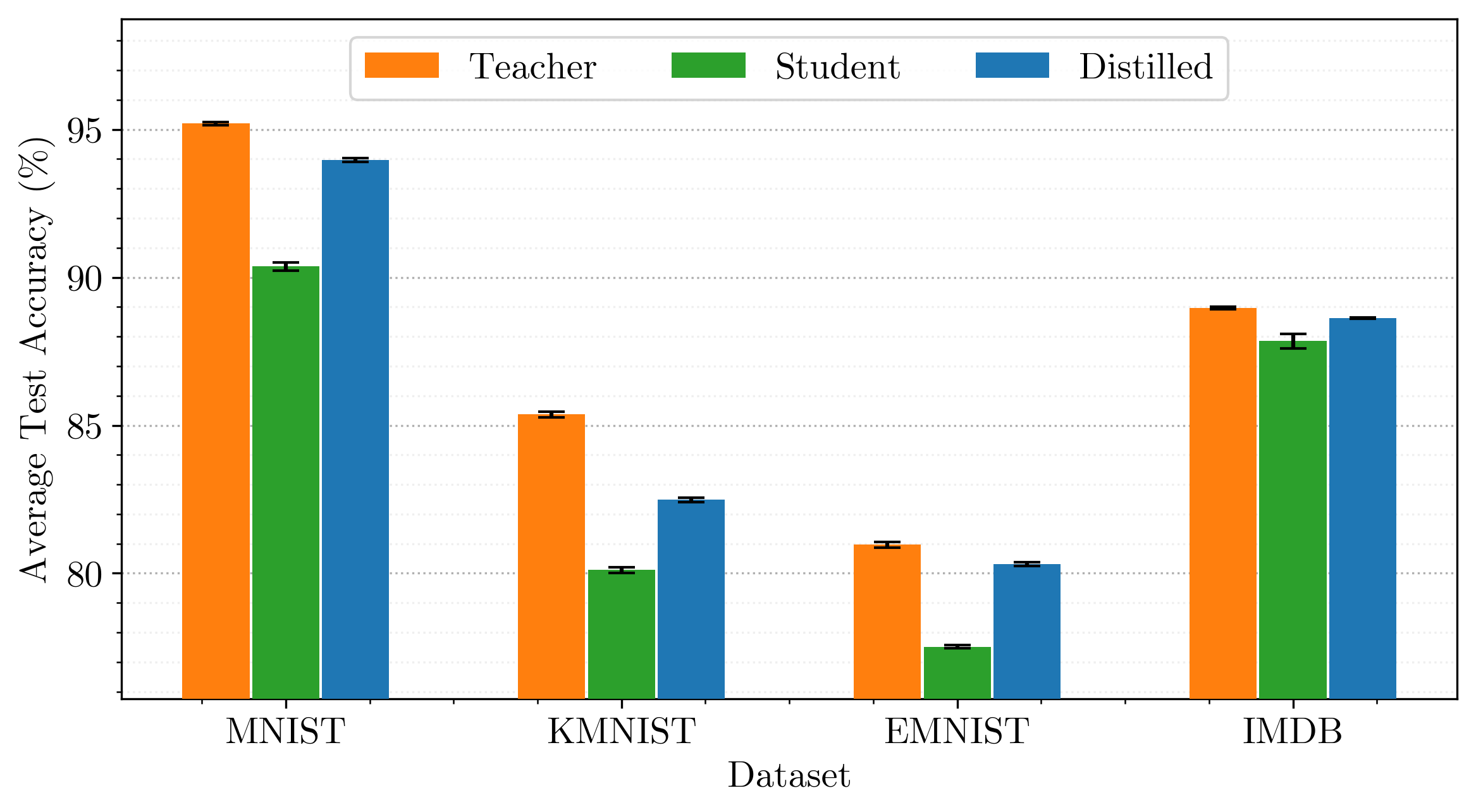}
    \caption{Average testing accuracy $Acc'$ on each dataset using DKD.}
    \label{fig:test-acc-combined}
\end{figure}
\begin{figure}
    \centering
    \includegraphics[width=\linewidth]{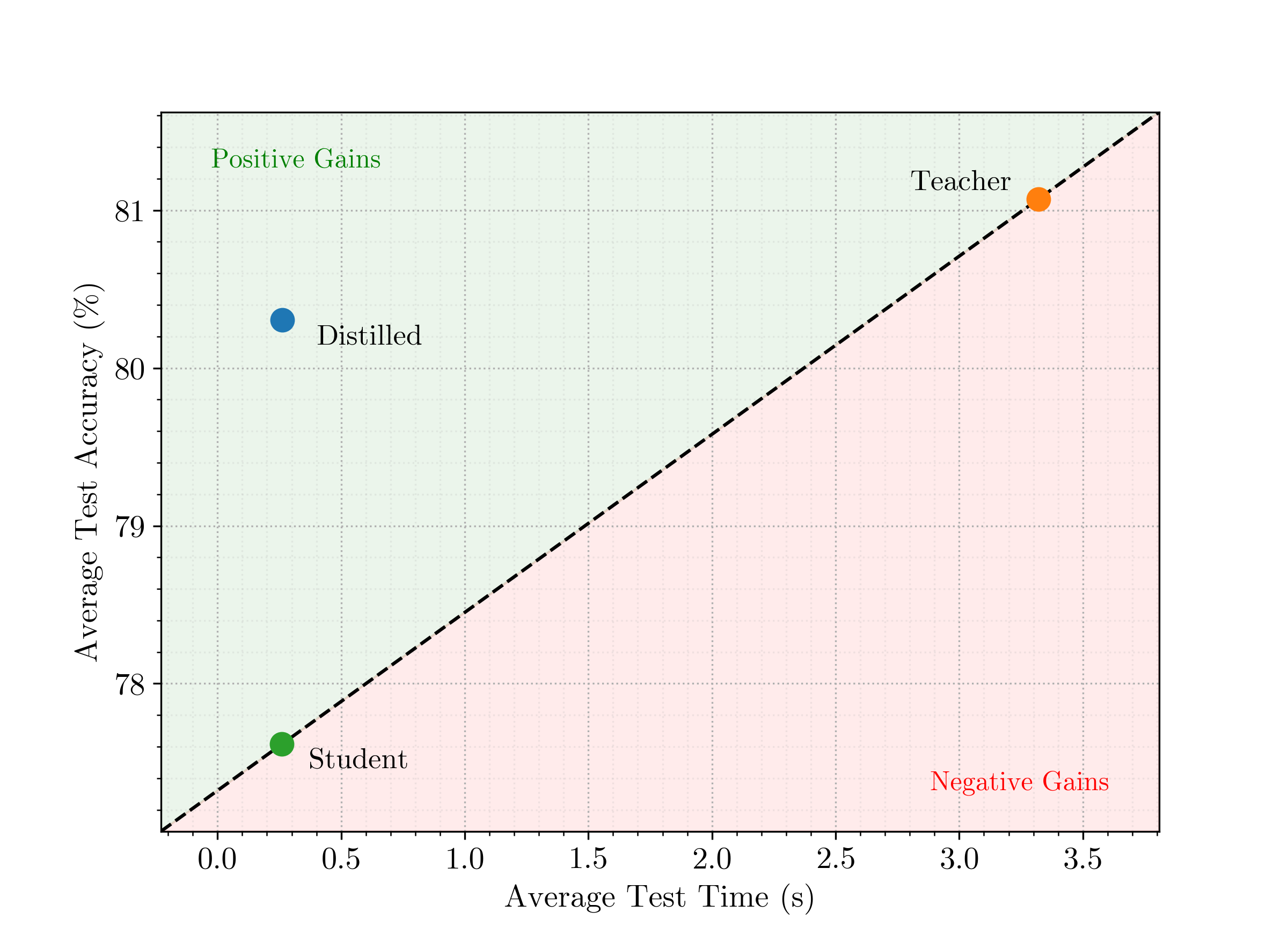}
        \caption{Testing efficiency on the EMNIST dataset using DKD.}
    \label{fig:test-eff-emnist}
\end{figure}

Figure~\ref{fig:test-acc-combined} shows the average accuracy of each model for each dataset. As described in Table~\ref{tab:test-table-dkd}, each distilled model significantly outperforms its respective baseline student model. 
\begin{figure}
    \centering
    \includegraphics[width=\linewidth]{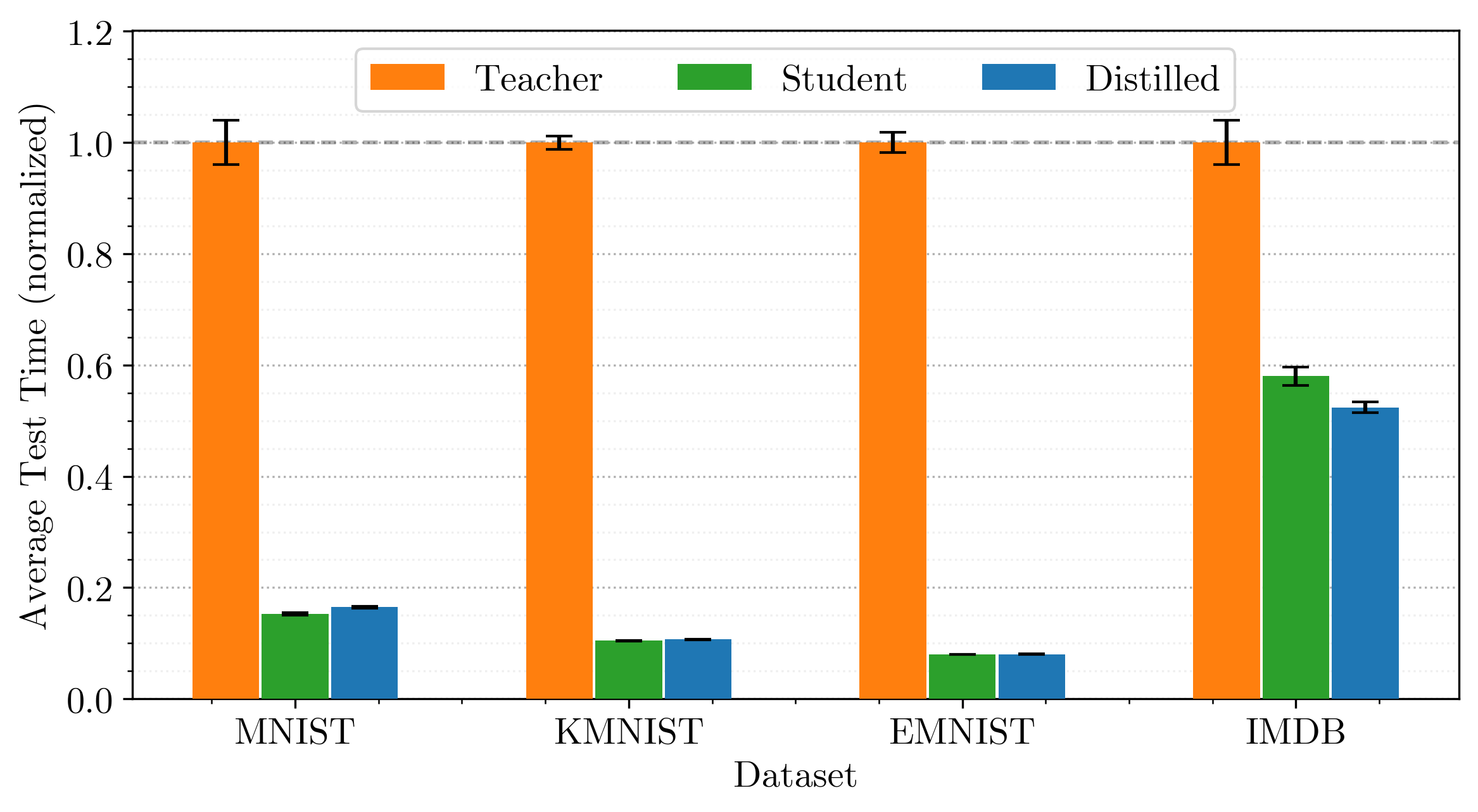}
    \caption{Average epoch testing time $\mathcal{T}'$ on each dataset using DKD (relative to teacher).}
    \label{fig:test-time-combined}
\end{figure}
Figure~\ref{fig:test-time-combined} shows the average testing time per epoch, normalized to the time of the teacher. As previously stated, the normalization step scales each dataset size to illustrate what the time would be if each dataset was the same size (see Figure~\ref{tab:dataset-size}). We see that each distilled model's average performance is roughly the same as its respective baseline student model. 
\begin{figure}
    \centering
    \includegraphics[width=\linewidth]{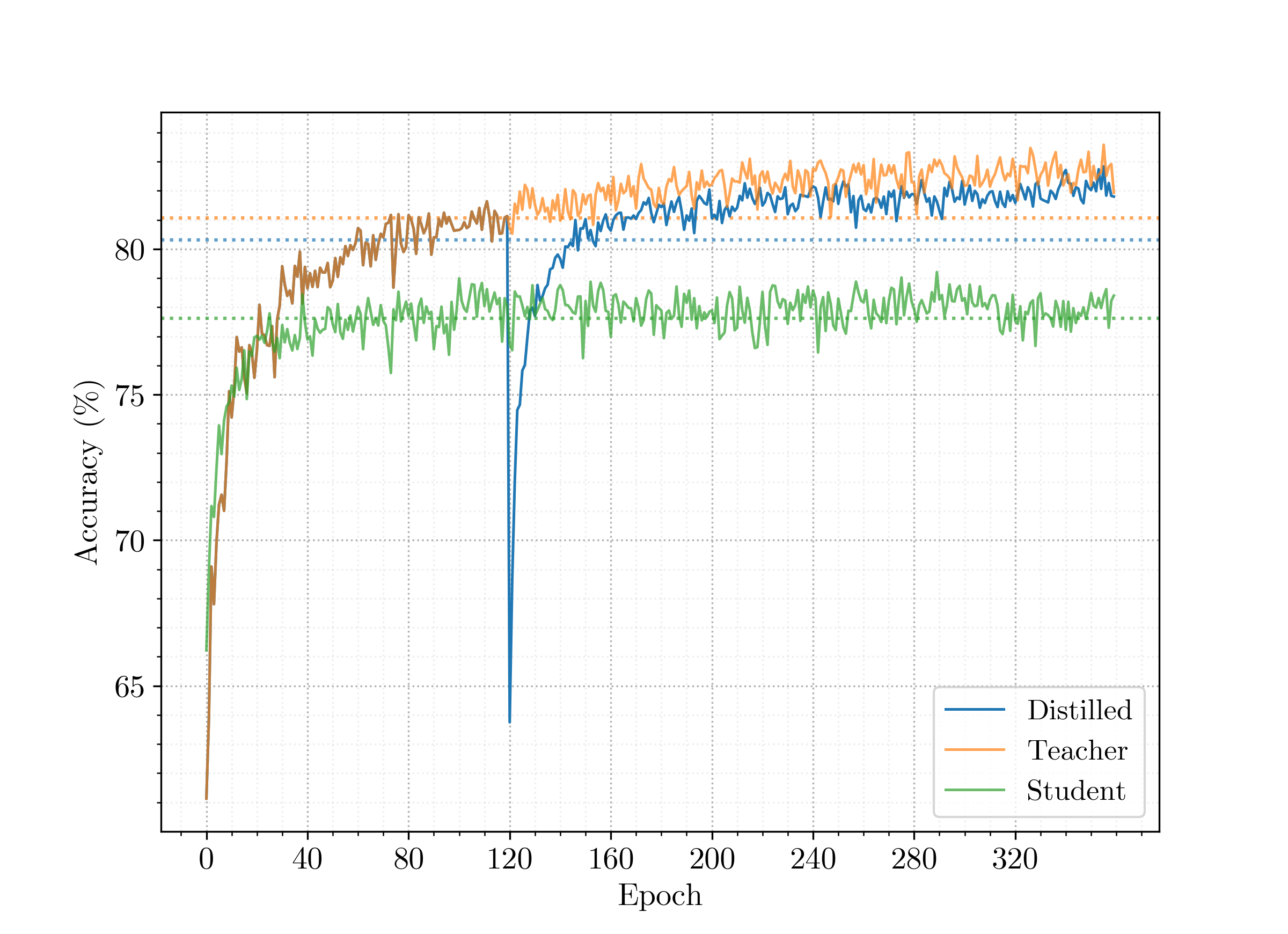}
    \caption{Testing accuracy $Acc$ on the EMNIST dataset using DKD.}
    \label{fig:test-acc-emnist}
\end{figure}
\begin{figure}
    \centering
    \includegraphics[width=\linewidth]{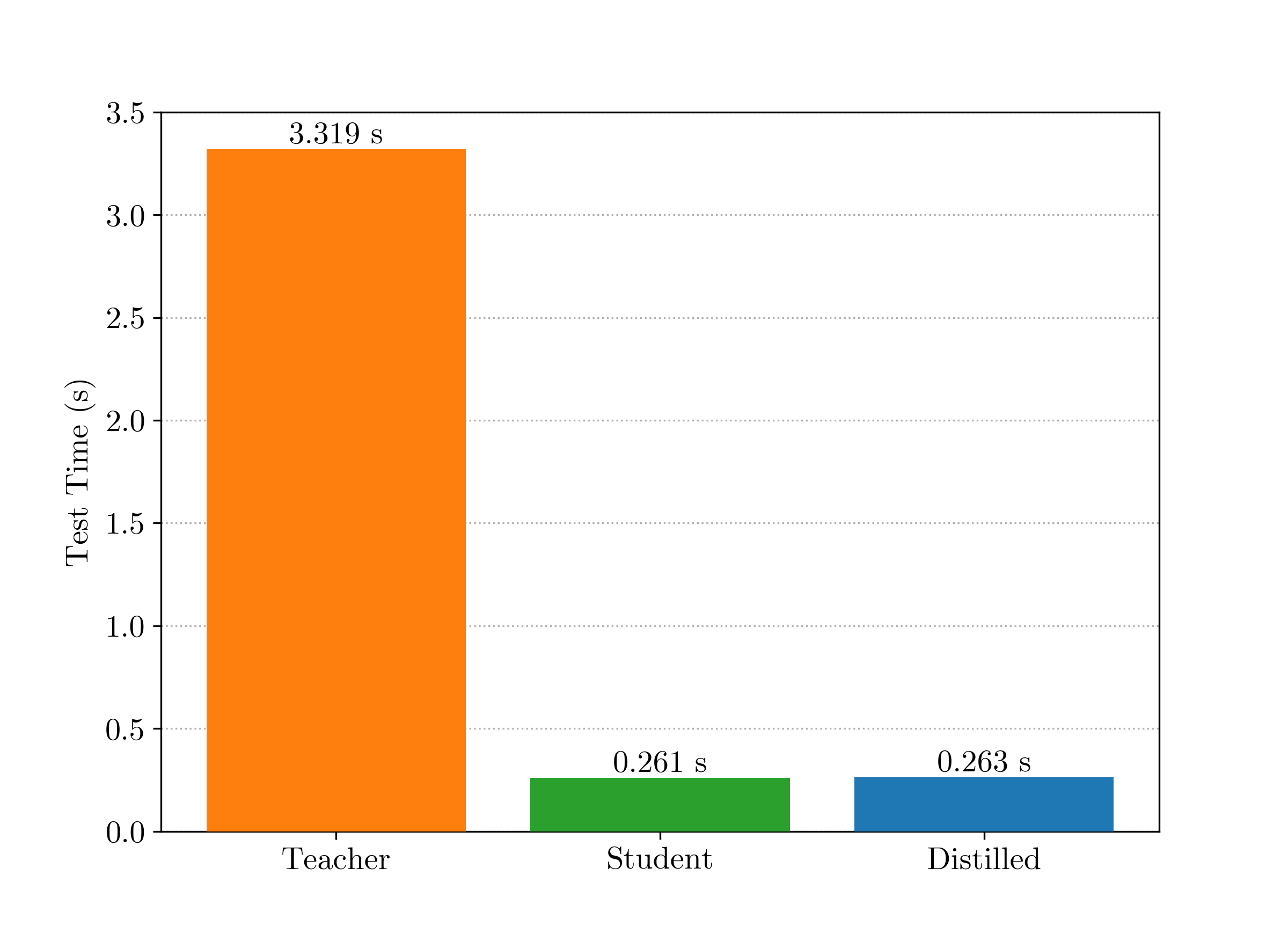}
    \caption{Average testing time $\mathcal{T}$ on the EMNIST dataset using DKD (relative to teacher).}
    \label{fig:test-time-emnist}
\end{figure}
The distilled model on the EMNIST dataset demonstrates incredible performance. As illustrated in Figure~\ref{fig:test-eff-emnist}, the accuracy of the distilled model almost reaches that of the teacher, but with a 12x reduction in inference time. The accuracy is comparable to multi-layer Tsetlin Machines described in~\cite{gorbenko2024multi}, but with significantly reduced inference time and memory usage. This accuracy and performance gain is visualized in Figures~\ref{fig:test-acc-emnist} and \ref{fig:test-time-emnist}.  

We observe similar performance gains in the MNIST dataset. The testing results mirror the training results in both accuracy in time, with a linearly downward shift in accuracy.

Knowledge distillation on the KMNIST testing dataset behaves in ways similar to the KMNIST training dataset. The distilled model accuracy beats the baseline student by 2.5 percentage points while maintaining the same execution time.

The text dataset IMDB also performs well. Accuracy for the distilled model is increased by 1 percentage point, and the inference time is decreased below both the baseline student and teacher models.

Note that for all models we observe a temporary drop in accuracy in the distilled model at $E_T$ epochs for the same reason described in section \ref{sec:trapa}.

\subsection{Activation Analysis}

\begin{figure}
    \centering
    \includegraphics[width=\linewidth]{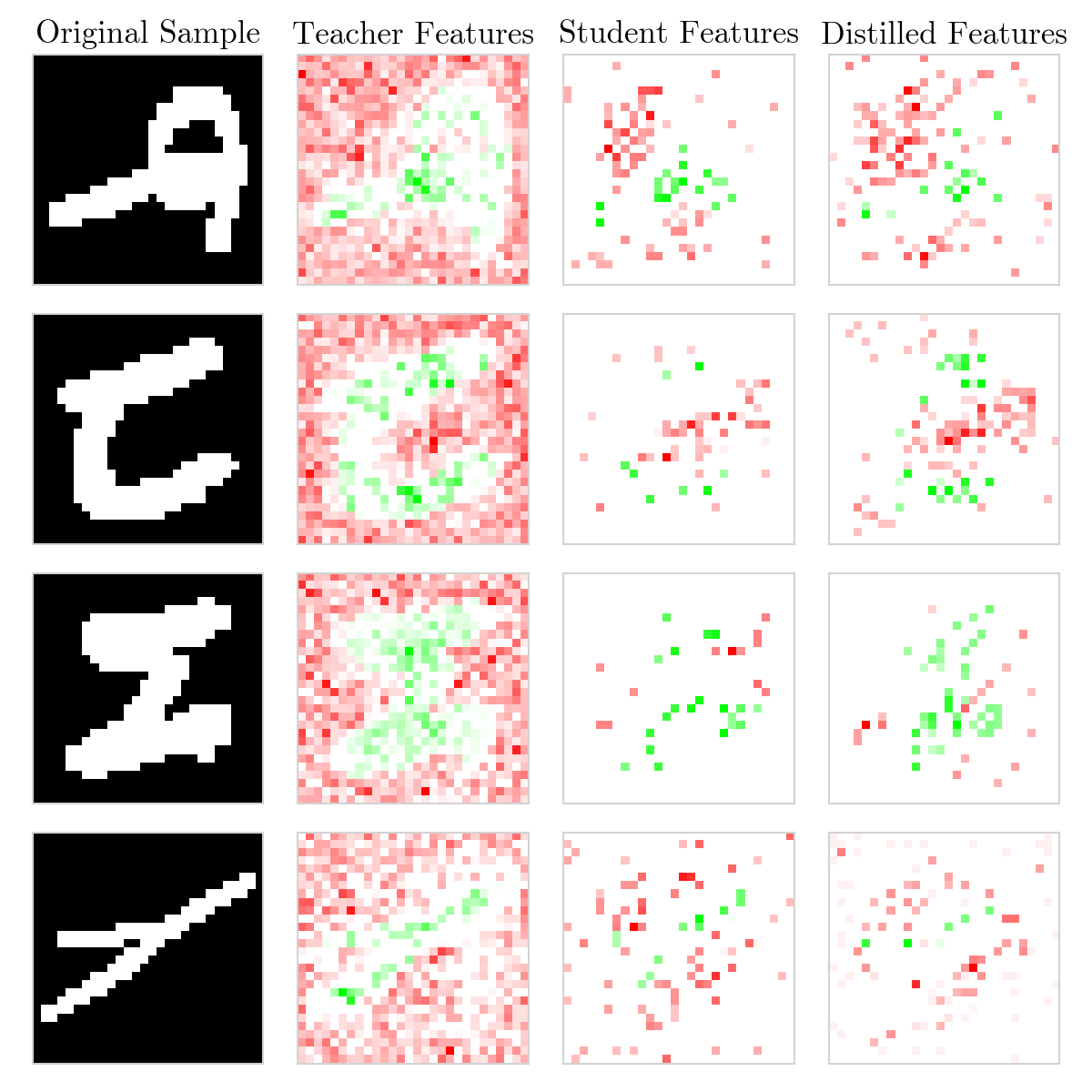}
    \caption{Activation maps for $TM_T$, $TM_S$, and $TM_D$ on 4 samples from the EMNIST dataset.}
    \label{fig:activation-emnist}
\end{figure}

Since each Tsetlin Machine takes the same input data, the impact of the teacher's influence on the student can be visualized through activation maps on the input data. Figure~\ref{fig:activation-emnist} shows an activation map generated on a sample of the testing data. Calculated over all TM clauses, green pixels signify included positive features (e.g., a white pixel); red pixels signify included negated features (e.g., a black pixel). The intensity of a pixel indicates how important that feature is to the TM.
Clearly the teacher activation maps are much denser than than the student models. This is unsurprising, given that the student model in this case has 10x fewer clauses than teacher. The distilled model, while having the same number of clauses as the teacher, is able to have a slightly wider range of included features. This is achieved from the extra context of the teacher's clause transfer and is what drives the increased accuracy in each distilled model.\footnote{Activation maps can be easily generated for image datasets, but are harder to visualize for textual datasets such as IMDB.}


\section{Comparison of Approaches}

Although both clause-based and distribution-enhanced knowledge distillation have valid use cases, our results suggest that DKD is the superior method in most scenarios. The central idea of knowledge distillation is to use a large model to enhance the performance of a smaller, memory-constrained model. The smaller model is usually applied in places where the larger model can't fit, such as in edge computing cases with limited resources. The teacher and student models can be trained on a powerful computer and then the student can be copied to the system with less resources for performing inference. That process is fully supported with our DKD algorithm. However, because the CKD algorithm must retain the teacher after training, CKD cannot be used in places where memory is restricted. As demonstrated in our experiments, CKD is more suited to applications where the user is looking only to a) reduce training time, or b) achieve a final accuracy greater than the teacher's.

Based on our results, we recommend our novel distribution-enhanced knowledge distillation as a feasible solution to implementing knowledge distillation in Tsetlin Machines.

\chapter{Conclusion}

This paper puts forward a novel framework for implementing knowledge distillation in Tsetlin Machines. It introduces a hybrid feature- and response-based method to distill information from a larger teacher model to a smaller student model. Our results have shown that a small distilled model can significantly outperform a parametrically identical model at the same size, increasing accuracy without increasing latency. This has consequential results in the relatively new domain of Tsetlin Machines. Notably, knowledge distillation was achieved without any sort of loss function, which is a key component in traditional, neural-network-based distillation. The exclusion of a loss function preserves the ability of Tsetlin Machines to run on low power, edge computing devices where complex math operations are prohibited or computationally expensive. 

Further research could be directed at optimizing the balance $\alpha$, weight transfer $z$, and temperature $\tau$ parameters, or exploring distillation from a neural network to Tsetlin Machine. Future work could also explore using different threshold $T$ and specificity $s$ parameters for the teacher and student Tsetlin Machines. The clause transfer algorithm could also be improved by considering clause coverage on all features, not just diversity and weight. Instead of solely using offline distillation as shown in this paper, further research could also be directed at comparing the effects of online and offline distillation.

\clearpage

\clearpage

\end{document}